\documentclass{article}

\usepackage[preprint]{neurips_2026}

\usepackage[utf8]{inputenc} % allow utf-8 input
\usepackage[T1]{fontenc}    % use 8-bit T1 fonts
\usepackage{hyperref}       % hyperlinks
\usepackage{url}            % simple URL typesetting
\usepackage{booktabs}       % professional-quality tables
\usepackage{amsfonts}       % blackboard math symbols
\usepackage{nicefrac}       % compact symbols for 1/2, etc.
\usepackage{microtype}      % microtypography
\usepackage{xcolor}         % colors
\usepackage{amsmath}
\usepackage{multirow}
\usepackage{booktabs}
\usepackage[table]{xcolor}
\usepackage{graphicx}
\usepackage{booktabs}
\usepackage[utf8]{inputenc}
\usepackage{booktabs}
\usepackage{array}
\usepackage{wrapfig}
\usepackage{lipsum}
\usepackage{tabularx}
\usepackage{amsmath}
\usepackage{enumitem}
\usepackage{bbding}

\definecolor{Blue}{RGB}{219,225,238}
\definecolor{Green}{RGB}{225,230,228}
\definecolor{Pink}{RGB}{252,241,250}

\title{LaST-R1: Reinforcing Robotic Manipulation via Adaptive Physical Latent Reasoning}

\author{
Hao Chen\thanks{Equal contribution, $^{\dagger}$Project lead, \textsuperscript{\Envelope}Corresponding author.} \hspace{0.1mm} \textsuperscript{\rm 1, 2} \hspace{0.1mm}
Jiaming Liu$^{*, \dagger}$\textsuperscript{\rm 2} \hspace{0.1mm}
Zhonghao Yan$^{*}$\textsuperscript{\rm 2} \hspace{0.1mm}
Nuowei Han$^{*}$\textsuperscript{\rm 2} \hspace{0.1mm}
Renrui Zhang$^{\dagger}$\textsuperscript{\rm 1} \hspace{0.1mm} \\
\textbf{Chenyang Gu\textsuperscript{\rm 2}} \hspace{0.1mm}
\textbf{Jialin Gao\textsuperscript{\rm 1}} \hspace{0.1mm}
\textbf{Ziyu Guo\textsuperscript{\rm 1}} \hspace{0.1mm}
\textbf{Siyuan Qian\textsuperscript{\rm 2}} \hspace{0.1mm}
\textbf{Yinxi Wang\textsuperscript{\rm 2}} \\
\textbf{Peng Jia\textsuperscript{\rm 3}} \hspace{0.1mm}
\textbf{Shanghang Zhang\textsuperscript{\rm 2}~\textsuperscript{\Envelope}} \hspace{0.1mm}
\textbf{Pheng-Ann Heng\textsuperscript{\rm 1}}
\vspace{0.2cm}\\
\textsuperscript{\rm 1}The Chinese University of Hong Kong 
\textsuperscript{\rm 2}State Key Laboratory of Multimedia Information \\Processing, School of Computer Science, Peking University
\textsuperscript{\rm 3}Simplexity Robotics \vspace{0.2cm}\\
Project page: \url{https://siriyep.github.io/last-r1/}
}

\begin{document}

\maketitle

%%%%%%%%%%%%%%%%%%%%%%%%%%%%%%%%%%%%%%%%%%%%%%%%%%%%%%%%%%%%

\begin{abstract}
Robotic foundation models require reasoning over complex visual scenes to execute adaptive actions in dynamic environments. 
While recent studies on latent-reasoning Vision-Language-Action (VLA) models have demonstrated the capability to capture fine-grained physical dynamics, they remain predominantly confined to static imitation learning, severely limiting their adaptability and generalization.
In this paper, we present \textbf{LaST-R1}, a novel reinforcement learning (RL) post-training framework designed to effectively harness ``latent reasoning-before-acting'' policies. Specifically, we propose \textbf{Latent-to-Action Policy Optimization (LAPO)}, a core RL algorithm that jointly optimizes the latent reasoning process and the action generation. By explicitly embedding latent Chain-of-Thought (CoT) reasoning directly within the RL optimization loop, LAPO stimulates profound physical world modeling, which in turn drives robust execution in interactive environments. Furthermore, an \textbf{adaptive latent CoT mechanism} is introduced, allowing the policy to dynamically modulate its reasoning horizon based on diverse environment states.
Experiments show that LaST-R1 achieves a near-perfect 99.9\% average success rate on the LIBERO benchmark with only one-shot supervised warm-up, significantly improving convergence speed and performance over prior state-of-the-art (SOTA) methods. In real-world deployments, LaST-R1 yields up to a 22.5\% average improvement over SOTA supervised fine-tuning approach across four complex tasks, including both single-arm and dual-arm settings. Finally, LaST-R1 demonstrates strong generalization across simulated and real-world environments.

\end{abstract}

\begin{figure*}[t]
    \centering
    \vspace{-0.2cm}
    \includegraphics[width=\textwidth]{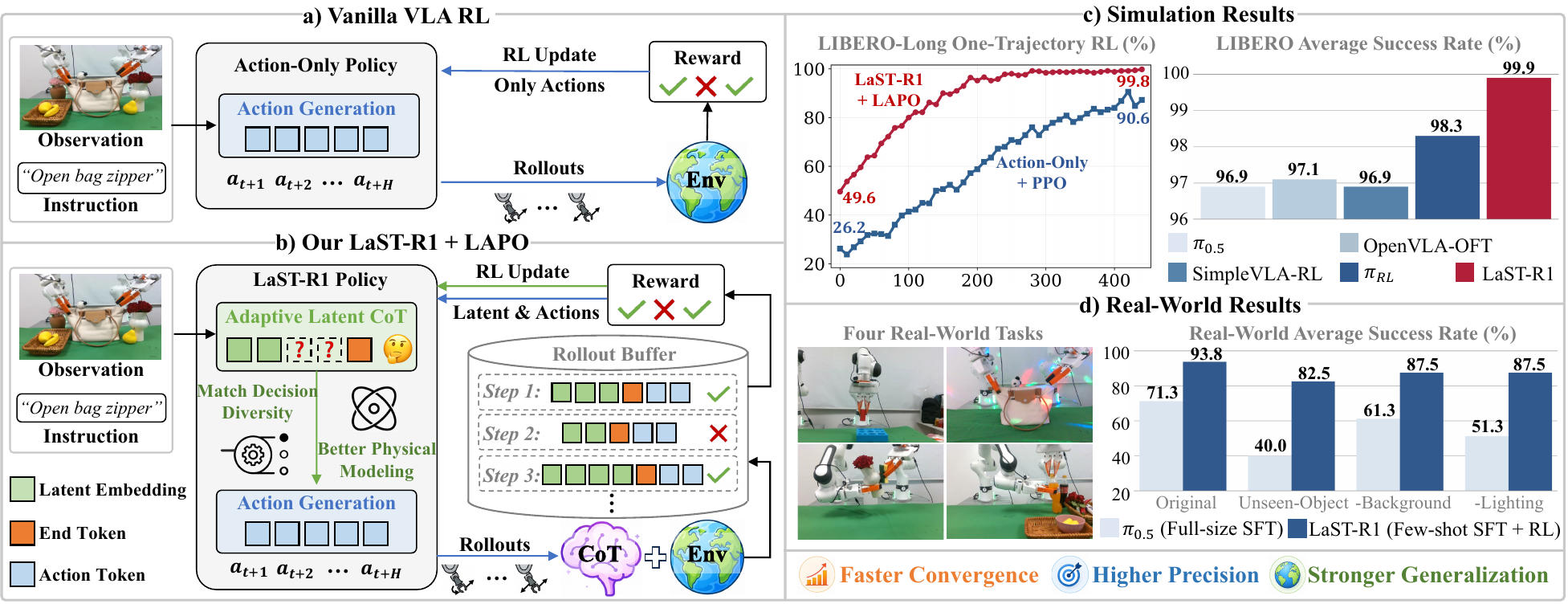} 
    \vspace{-0.5cm}
    \caption{\textbf{The Overview of LaST-R1.} (a) Unlike vanilla RL baselines that strictly optimize actions, (b) our approach utilizes LAPO to jointly optimize an adaptive latent CoT alongside physical execution. By bridging cognitive reasoning and control, LaST-R1 achieves (c) faster convergence speed, higher success rate in simulation, and (d) stronger generalization capabilities in real-world.}
    \label{fig:teaser}
    \vspace{-0.2cm}
\end{figure*}

%%%%%%%%%%%%%%%%%%%%%%%%%%%%%%%%%%%%%%%%%%%%%%%%%%%%%%%%%%%%

\section{Introduction}

Driven by large-scale pre-trained VLMs \cite{karamcheti2024prismatic,bai2025qwen3vltechnicalreport,beyer2024paligemma}, Vision-Language-Action (VLA) models \cite{kim2024openvla,kim2025fine,black2024pi_0,intelligence2025pi_,liu2025hybridvla,chen2025fast,li2024cogact} have emerged as a promising paradigm for general robotic manipulation.
Moving beyond direct observation-to-action mapping, recent architectures increasingly draw inspiration from Chain-of-Thought (CoT) reasoning \cite{wei2022chain}.
While explicitly generating linguistic traces \cite{ye2025vla,huang2025thinkact,lin2025onetwovla,zawalski2024robotic} or future states \cite{zhao2025cot,gu2025manualvla,liu2025mla,cen2025worldvla,intelligence2026pi} provides structured guidance before acting, it incurs non-negligible inference latency and discretization bottlenecks.
This inherently restricts the model's ability to capture continuous, high-frequency physical dynamics.
Therefore, recent research \cite{liu2026last,cai2026internvla} has pivoted toward reasoning in a compact latent space that accommodates fine-grained, hard-to-verbalize physical knowledge, establishing a highly expressive reasoning foundation for complex manipulation.

Despite the advantages of latent reasoning, existing frameworks remain grounded in imitation learning, requiring massive, high-cost expert demonstrations.
Their reliance on static datasets prevents closed-loop environmental interaction, leading to compounding errors and limited generalization.
Concurrently, a nascent line of research \cite{chen2025pirl,li2025simplevla,lu2025vla,chen2025conrft,xu2026twinrl,liu2025can,li2025gr,intelligence2025pi} has introduced online reinforcement learning (RL) for VLA post-training, improving exploration and robustness through trial-and-error interaction with the environment.
However, as shown in Figure \ref{fig:teaser} (a), current methods are largely restricted to vanilla architectures that operate directly in the action space, bypassing the underlying physical reasoning process.
Consequently, they lack interaction-driven learning to accurately model physical dynamics, limiting their ability to adapt to complex environments.

To this end, we propose \textbf{LaST-R1}, a novel RL post-training framework specifically developed to optimize ``latent reasoning-before-acting'' policies, as shown in Figure \ref{fig:teaser} (b).
We first present the LaST-R1 VLA model, which tightly integrates latent CoT reasoning with action generation. Specifically, the model autoregressively produces latent reasoning embeddings to capture structured physical dynamics, serving as conditions for parallel action decoding. To ensure stable and expressive reasoning, these latent embeddings are grounded in global feature representations from a vision foundation model~\cite{simeoni2025dinov3}, providing a robust semantic and spatial prior.
Building on this architecture, we propose \textbf{Latent-to-Action Policy Optimization (LAPO)}, a RL algorithm that jointly optimizes the latent reasoning process and action generation.
Unlike prior RL approaches that operate solely within the action space \cite{li2025simplevla,chen2025pirl,lu2025vla}, LAPO treats latent reasoning embeddings as implicit decision variables. By bridging this implicit deliberation with action generation through a unified step-level likelihood ratio, our method allows environmental rewards to directly shape the latent reasoning space, thereby concurrently optimizing the model's physical comprehension and interactive robustness.
Finally, an \textbf{adaptive latent CoT mechanism} is introduced to augment the framework, enabling the dynamic adjustment of the reasoning horizon based on task diversity. This allows the policy to allocate more reasoning steps for intricate manipulations, while maintaining fast execution for reactive responses, effectively balancing cognitive capacity and computational efficiency.

To empirically validate our framework, LaST-R1 VLA is first initialized with pre-trained Qwen3-VL-4B \cite{bai2025qwen3vltechnicalreport} and undergoes large-scale pre-training across diverse robotic manipulation datasets \cite{o2024open,wu2024robomind,khazatsky2024droid}. For specific downstream tasks, the policy is adapted through a supervised fine-tuning (SFT) warm-up before engaging in online RL post-training.
During evaluation, our approach achieves a near-perfect 99.9\% average success rate on the LIBERO benchmark with one-shot warmup, outperforming prior state-of-the-art (SOTA) baselines, with a particularly notable improvement of 5.3\% on the LIBERO-Long suite (long-horizon tasks).
In real-world deployments, LaST-R1 yields up to a 22.5\% improvement over the SOTA SFT method across four complex tasks, reaching a 90\% average success rate in both single-arm and dual-arm settings.
Furthermore, systematic evaluations demonstrate that the proposed framework exhibits strong generalization in simulation environments, consistently outperforming vanilla action-only RL policies trained with standard PPO \cite{schulman2017proximal}.
Notably, LaST-R1 achieves zero-shot generalization to unseen objects, backgrounds, and lighting conditions after LAPO post-training.  
In summary, our main contributions are as follows:

\begin{enumerate}[leftmargin=15pt, labelsep=5pt, label=\arabic*)]

\item We present \textbf{LaST-R1}, a specialized RL post-training framework for ``latent reasoning-before-acting'' policies to enhance physical world modeling and robust action generation.
\item We propose \textbf{Latent-to-Action Policy Optimization (LAPO)}, a novel RL algorithm that jointly optimizes latent reasoning and action generation via a unified step-level likelihood ratio.

\item We design an \textbf{adaptive latent CoT mechanism} that dynamically adjusts the reasoning horizon based on task diversity, balancing reasoning capacity and computational efficiency.

\end{enumerate}

\section{Methodology}

This section presents the proposed \textbf{LaST-R1} framework, as shown in Figure \ref{fig:method}. We first review the problem formulation (Section \ref{sec:prelimary}), then introduce a unified VLA architecture with latent Chain-of-Thought (CoT) reasoning (Section \ref{method:architecture}). Next, we present \textbf{Latent-to-Action Policy Optimization (LAPO)}, which jointly optimizes latent reasoning and action generation (Section \ref{sec:lapo}). Finally, we describe an \textbf{adaptive mechanism} that dynamically adjusts the reasoning horizon (Section \ref{sec:adapt_cot}).

\subsection{Preliminaries}
\label{sec:prelimary}
We formulate the Vision-Language-Action (VLA) model as a parameterized policy $\pi_\theta$ that maps multimodal observations (i.e., visual inputs and language instructions) to an action chunk $\mathbf{a}_{t:t+H}$ in $SE(3)$ space. For single-arm robots such as the Franka Research 3, the action is a 7-DoF end-effector control vector $\mathbf{a}_t \in \mathbb{R}^7$, comprising 3-DoF positional offsets, a 3-DoF orientation (represented as Euler angles), and a 1-DoF gripper state. For dual-arm systems, we extend this formulation to 14-DoF via vector concatenation. The policy can be optimized either through supervised fine-tuning (SFT) using expert demonstrations or via reinforcement learning (RL) via environment interaction.

\textbf{SFT Formulation.}
In supervised fine-tuning, we have access to a dataset of expert demonstrations $\mathcal{D} = \{(\mathbf{s}_t, \mathbf{a}_{t:t+H})\}$, where $\mathbf{s}_t$ denotes the multimodal observations at timestep $t$. The policy is trained to imitate expert behavior by maximizing the conditional log-likelihood:
\begin{equation}
\mathcal{J}_{\text{SFT}}(\theta) = \mathbb{E}_{(\mathbf{s}_t, \mathbf{a}_{t:t+H}) \sim \mathcal{D}} \left[ \log \pi_\theta(\mathbf{a}_{t:t+H} \mid \mathbf{s}_t) \right].
\end{equation}
While this objective enables effective behavioral cloning, the model's capability and generalization remain inherently bounded by the quality, scale, and diversity of the training data.

\textbf{RL Formulation.}
To surpass the limitations of static imitation, RL casts the sequential decision-making problem as an interactive process where the policy engages with the environment over trajectories of length $T+1$. At each timestep $t$, the agent observes a state $\mathbf{s}_t$ and samples an action chunk $\mathbf{a}_{t:t+H} \sim \pi_\theta(\cdot \mid \mathbf{s}_t)$. The environment then transitions to the next state $\mathbf{s}_{t+1}$ and yields a scalar reward $r_t$. This interaction induces a trajectory $\tau$, whose distribution is governed by the initial state and the environment dynamics. The objective is to maximize the expected discounted return:
\begin{equation}
\begin{aligned}
\mathcal{J}_{\text{RL}}(\theta) &= \mathbb{E}_{\tau \sim \pi_\theta} \left[ \sum_{t=0}^{T} \gamma^t r_t \right], \quad
\nabla_\theta \mathcal{J}_{\text{RL}}(\theta) = \mathbb{E}_{\tau \sim \pi_\theta} \left[ \sum_{t=0}^{T} \nabla_\theta \log \pi_\theta(\mathbf{a}_{t:t+H} \mid \mathbf{s}_t) \, \hat{A}_t \right],
\end{aligned}
\end{equation}
where $\gamma \in [0, 1)$ is the discount factor, and $\hat{A}_t$ denotes the advantage estimate, capturing the relative utility of the sampled action chunk $\mathbf{a}_{t:t+H}$ compared to the expected value at state $\mathbf{s}_t$.

\subsection{LaST-R1 Model Architecture}
\label{method:architecture}

\begin{figure*}[t]
    \centering
    \vspace{-0.1cm}
    \includegraphics[width=\textwidth]{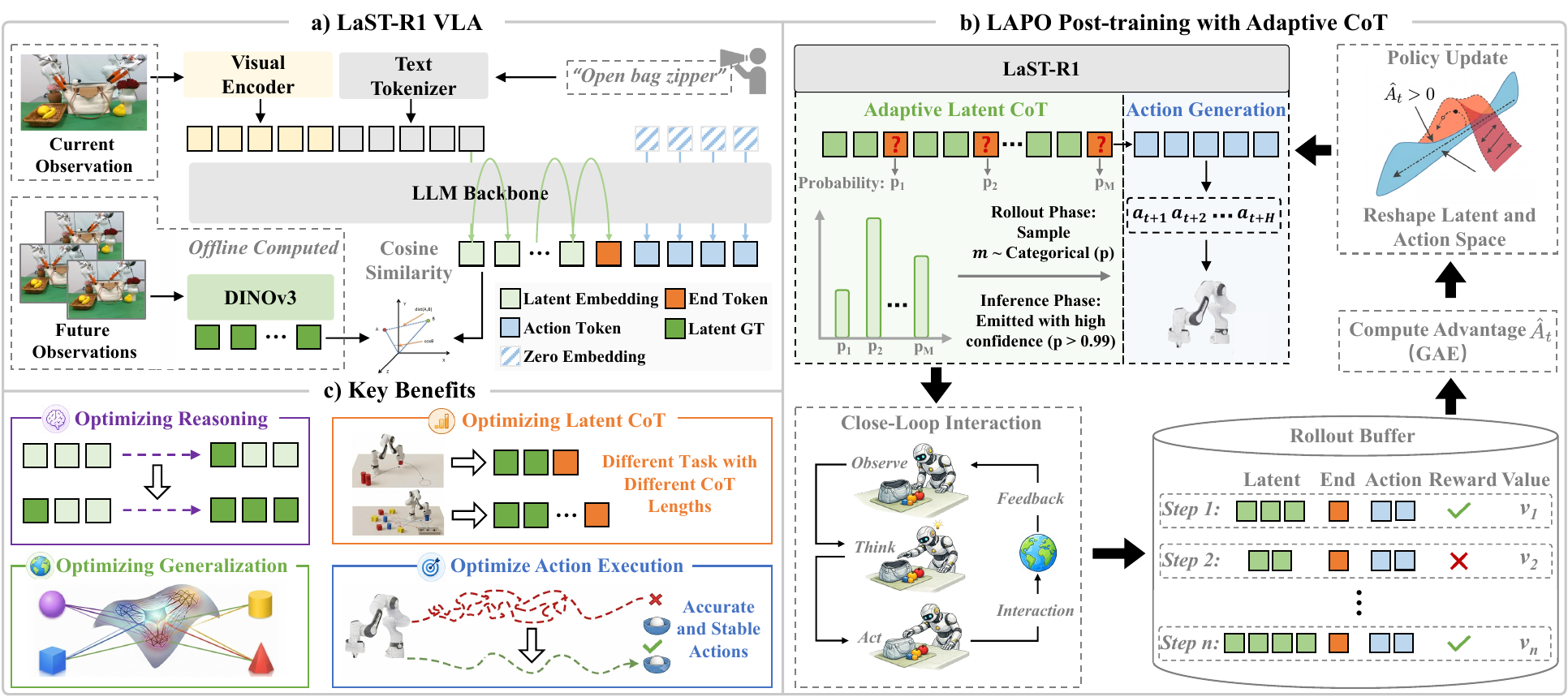} 
    \vspace{-0.6cm}
    \caption{\textbf{The Framework of LaST-R1.} 
    (a) LaST-R1 VLA is a unified model that takes visual observations and language instructions as input, where a vision foundation model provides physically grounded latent targets to guide latent CoT reasoning before action generation.
    (b) During LAPO RL post-training, LaST-R1 interacts with the environment in a closed loop manner, storing latents, actions, and rewards in a rollout buffer for jointly reshaping the latent and action spaces. It further enables adaptive reasoning by learning to emit the \texttt{<latent\_end>} token based on predicted probabilities, dynamically adjusting the reasoning horizon across tasks.
    (c) Through LAPO, LaST-R1 achieves adaptive reasoning lengths across diverse tasks, improving generalization and execution stability.
    }
    \label{fig:method}
    \vspace{-0.3cm}
\end{figure*}

As shown in Figure \ref{fig:method} (a), we propose a unified LaST-R1 VLA model that jointly models latent reasoning and action generation. We first present the overall pipeline, followed by a novel method for constructing compact latent representations of future dynamics.

\noindent \textbf{Overall Pipeline.}
LaST-R1 VLA builds upon the Qwen3-VL-4B~\cite{bai2025qwen3vltechnicalreport} architecture, integrating a SigLIP2-Large \cite{tschannen2025siglip} visual encoder with an LLM backbone. Given an input image $I \in \mathbb{R}^{H \times W \times 3}$, the visual encoder first employs 2D-RoPE with interpolated absolute positional embeddings to transform spatial information into dense visual embeddings $f_{\text{v}} \in \mathbb{R}^{ N_{\text{v}} \times 2560}$. Subsequently, these embeddings are concatenated with language embeddings $f_{\text{l}} \in \mathbb{R}^{ N_{\text{l}} \times 2560}$ and processed by the LLM backbone, where $N_{\text{v}}$ and $N_{\text{l}}$ denote the respective sequence lengths.
The model first generates latent reasoning embeddings, followed by action prediction.
To fully leverage the strong reasoning capability of the LLM, latent embeddings are generated in an autoregressive manner, with their representation detailed in the next paragraph. To support action generation, we normalize and discretize continuous actions into tokens, extending the LLM vocabulary accordingly~\cite{kim2024openvla}. We adopt a parameter-free action tokenizer to map between continuous actions and discrete tokens, while employing parallel decoding to improve inference efficiency~\cite{kim2025fine}.
To enable RL post-training, we introduce a value head composed of a 4-layer MLP to estimate state values, which shares the same backbone with the VLA~\cite{liu2025can,chen2025pirl}.

\noindent \textbf{Latent Representation.}
To model temporal environmental dynamics, LaST-R1 VLA autoregressively infers future states over a horizon $N_z$, yielding $N_z$ latent reasoning embeddings. Prior methods extract these latents via average pooling~\cite{liu2026last} or auxiliary learnable parameters~\cite{cai2026internvla}, which often sacrifices fine-grained spatial structures or injects undesired inductive biases into the learned latent space. To overcome this, we construct latent embeddings using DINOv3~\cite{simeoni2025dinov3}, a state-of-the-art vision foundation model. 
Specifically, we extract its \texttt{<CLS>} token ($f_{\text{d}} \in \mathbb{R}^{ 1 \times 4096}$) as a holistic image embedding and apply top-$k$ ($k$ = 2560) selection based on feature magnitude along the channel dimension.
These selected latent targets align with the hidden size of the VLA embedding while preserving the most salient visual components.
By harnessing DINOv3's structurally rich and semantically dense feature space, the extracted \texttt{<CLS>} token serves as a strong semantic prior, facilitating accurate modeling of physical dynamics.
Crucially, these latent targets are precomputed offline, effectively decoupling the foundation model from the policy to ensure no additional overhead during both training and inference.
We systematically compare this approach with prior latent representation formats in Section~\ref{exp:ae} to further validate its effectiveness.

\subsection{Latent-to-Action Policy Optimization}
\label{sec:lapo}
To effectively train the proposed \textit{latent reasoning-before-acting} policy LaST-R1 VLA, we introduce \textbf{Latent-to-Action Policy Optimization (LAPO)}, a RL framework that jointly optimizes latent reasoning and action generation, as shown in Figure \ref{fig:method} (b). In our formulation, the policy first generates latent embeddings, which condition subsequent action prediction. 
While prior works \cite{liu2025can, zang2025rlinf} have established PPO \cite{schulman2017proximal} as an effective post-training method for VLA models, they operate solely in the action space and overlook intermediate reasoning processes that are integral to decision-making.
To address this limitation, we propose LAPO, which treats latent embeddings as \emph{implicit decision variables} and introduces a unified optimization framework that enables environment reward signals to shape both reasoning and action generation spaces across the entire trajectory.

\noindent \textbf{Rollout Collection.}
At each environment step $t$, given the current state $\mathbf{s}_t$ comprising visual observations and language instructions, the policy first generates a sequence of latent embeddings $\mathbf{Z}_t^{\text{old}} = \{\mathbf{z}_{t,i}^{\text{old}}\}_{i=1}^{N_z}$ autoregressively: $\mathbf{z}_{t,i}^{\text{old}} \sim \pi_\theta(\cdot \mid \mathbf{s}_t, \mathbf{z}_{t,<i}^{\text{old}})$. Conditioned on these latents, a sequence of discrete action tokens $\mathbf{C}_t = \{\mathbf{a}_{t,j}\}_{j=1}^{N_a}$ is then produced, where $N_a$ denotes the total number of tokens representing the $H$-step action chunk.
Specifically, we reuse the KV cache from the latent generation phase and decode the action tokens in parallel via a single forward pass with bidirectional attention over $N_a$ placeholder vectors.
We further insert a special \texttt{<latent\_end>} token between the latent and action sequences to signify the completion of reasoning. The final hidden embedding of this token is routed into a value head to output the state value estimation $v_t$. To formulate a unified action distribution for decision step $t$, we compute the joint log-probability of the action sequence by summing over the individual tokens: 
$\log \pi_\theta(\mathbf{C}_t \mid \cdot) = \sum_{j=1}^{N_a} \log \pi_\theta(\mathbf{a}_{t,j} \mid \cdot)$.
During rollout, we store the sampled latent sequences $\mathbf{Z}_t^{\text{old}}$, action token sequences $\mathbf{C}_t$, their corresponding log-probabilities, and the estimated state values. Ultimately, a scalar reward is assigned based on task success or failure, yielding complete trajectories that are subsequently used to compute the advantage estimates $\hat{A}_t$ via Generalized Advantage Estimation (GAE) for policy optimization.

\noindent \textbf{Policy Optimization.}
During policy updates, the model processes the identical rollout context to yield updated continuous latent embeddings $\mathbf{Z}_t^\theta = \{\mathbf{z}_{t,i}^\theta\}_{i=1}^{N_z}$ and updated discrete action log-probabilities. Because the advantage $\hat{A}_t$ is computed at the decision-step level, we define a step-level ratio $r_t(\theta)$ for both the action sequence and the continuous latents.
For the discrete action sequence, the ratio is computed via the joint probabilities: $r_t^a(\theta) = \exp \left( \log \pi_\theta(\mathbf{C}_t \mid \cdot) - \log \pi_{\theta_{\text{old}}}(\mathbf{C}_t \mid \cdot) \right)$.
For the continuous latent embeddings, we approximate the distribution of rollout latents $\mathbf{Z}_t^{\text{old}}$ using an isotropic Gaussian centered at the current policy output $\mathbf{Z}_t^\theta$.
By summing the squared Euclidean distances over the entire latent sequence length $N_z$, we define the step-level latent ratio surrogate as:
\begin{equation}
    r_t^{z}(\theta) = \frac{\pi_{\theta}(\mathbf{Z}_t^{\text{old}} \mid \cdot)}{\pi_{\theta_{\text{old}}}(\mathbf{Z}_t^{\text{old}} \mid \cdot)} = \exp\left(-\frac{1}{2\sigma^2} \sum_{i=1}^{N_z} \| \mathbf{z}_{t,i}^{\text{old}} - \mathbf{z}_{t,i}^{\theta} \|^2 \right),
\end{equation}
where $\sigma$ is a fixed hyperparameter regulating the latent variance. For a more intuitive, step-by-step derivation of $r_t^{z}(\theta)$, please refer to Appendix \ref{appendix:latent_likelihood_ratio}.
By jointly optimizing reasoning latents and actions, we formulate the joint LAPO clipped surrogate loss as an expectation over all valid decision steps $t$:
\begin{equation}
    \mathcal{L}_{\text{policy}}(\theta) = - \mathbb{E}_t \left[ \sum_{k \in \{z, a\}} \min \left( r_t^k(\theta) \hat{A}_t, \text{clip}(r_t^k(\theta), 1-\epsilon_{\text{min}}, 1+\epsilon_{\text{max}}) \hat{A}_t \right) \right],
\end{equation}
where $\epsilon_{\text{min}}$ and $\epsilon_{\text{max}}$ are the clipping thresholds. 
This objective enables the environment reward to shape both the action space and the internal reasoning space simultaneously. Specifically, when $\hat{A}_t > 0$, the optimization explicitly minimizes the latent distance $\sum_{i=1}^{N_z} \| \mathbf{z}_{t,i}^{\text{old}} - \mathbf{z}_{t,i}^{\theta} \|^2$, effectively pulling the current policy's latent representations toward the "good-reasoning" manifolds that facilitated successful trajectories.
In practice, to achieve better optimization flexibility, we decouple the joint policy objective $\mathcal{L}_{\text{policy}}(\theta)$ into two separate components: the action loss $\mathcal{L}_{\text{action}}(\theta)$ (for $k=a$) and the latent loss $\mathcal{L}_{\text{latent}}(\theta)$ (for $k=z$). Finally, the total training loss unifies these policy objectives with state-value estimation, governed by weighting coefficients $\lambda_1$ and $\lambda_2$:
\begin{equation}
    \mathcal{L}_{\text{total}}(\theta) = \mathcal{L}_{\text{action}}(\theta) + \lambda_1\mathcal{L}_{\text{latent}}(\theta) + \lambda_2 \mathcal{L}_{\text{value}}(\theta),
\end{equation}
where $\mathcal{L}_{\text{value}}(\theta) = \mathbb{E}_t \left[ (v_t - \hat{R}_t)^2 \right]$ is the mean squared error (MSE) against the target return $\hat{R}_t$.

\subsection{Adaptive Latent CoT Reasoning}
\label{sec:adapt_cot}
While fixed-length reasoning provides a baseline, it imposes a static cognitive horizon that fails to capture the inherent diversity of robotic tasks. Enforcing a fixed latent length $N_z$ incurs unnecessary computational overhead on highly predictable motions and restricts the reasoning capacity required for maneuvers demanding intricate cognitive planning.
To address this, we introduce an \textbf{adaptive latent CoT mechanism} optimized via our proposed RL framework. 
By allowing the model to adapt its reasoning length based on reward signals, we enable an ``early-exit'' strategy that dynamically tailors the latent reasoning horizon to the specific demands of each task.

\noindent \textbf{Dynamic Latent Generation.}
To enable adaptive reasoning, we shift the \texttt{<latent\_end>} token from a deterministic sequence terminator to a dynamically emitted transition signal within the latent generation process. During autoregressive generation, if the policy predicts the \texttt{<latent\_end>} token with a sufficiently high confidence probability $p \ge 0.99$ (where $p \in [0, 1]$), indicating that the model is highly certain the reasoning phase should conclude at the current decision step, the latent generation terminates and the model transitions to action prediction. 
To stabilize the training process and prevent erratic fluctuations in latent reasoning length, we restrict the emission of the \texttt{<latent\_end>} token to a predefined set of $M$ candidate positions, bounded by a maximum reasoning length $N_{\text{max}}$.

\noindent \textbf{Exploration via Length Sampling.}
During RL rollouts, exploring various reasoning lengths is crucial for discovering the optimal cognitive horizon across diverse environmental states.
We treat the reasoning length as a stochastic decision to encourage environmental interaction.
Specifically, we extract the pre-softmax logits $l$ associated with the \texttt{<latent\_end>} token across all $M$ predefined candidate positions, and then apply a temperature parameter $\beta$ to control the exploration entropy, yielding a normalized categorical distribution over the candidate lengths:
\begin{equation}
p_m = \frac{\exp(l_m / \beta)}{\sum_{i=1}^M \exp(l_i / \beta)}, \quad \forall m \in \{1, \dots, M\}.
\end{equation}
The reasoning length index $m \sim \texttt{Categorical}(p_1, \dots, p_M)$ is sampled to dictate the number of generated latent embeddings, facilitating exploration during rollout to discover adaptive cognitive horizons. During inference, the model adopts a confidence-based exit strategy for efficient execution.

\noindent \textbf{Adaptive Length Optimization.}
To optimize the adaptive length selection, the decision of \textbf{\emph{when}} to stop reasoning must be explicitly supervised. We introduce an additional policy loss term for the \texttt{<latent\_end>} token. 
Let $\mathbf{z}_t^{\text{end}}$ denote the transition token emitted at the sampled position $m$. We record its log-probability $\log \pi_\theta(\mathbf{z}_t^{\text{end}} \mid \cdot)$ during rollout. During the policy update, we construct a discrete likelihood ratio for the transition token:
$r_t^{\text{end}}(\theta) = \exp \left( \log \pi_\theta(\mathbf{z}_t^{\text{end}} \mid \cdot) - \log \pi_{\theta_{\text{old}}}(\mathbf{z}_t^{\text{end}} \mid \cdot) \right).$
For models equipped with adaptive reasoning, the total training objective is augmented to include this transition-specific term $\mathcal{L}_{\text{end}}(\theta)$, weighted by a coefficient $\lambda_3$:
\begin{equation}
\mathcal{L}_{\text{total}}(\theta) = \mathcal{L}_{\text{action}}(\theta) + \lambda_1 \mathcal{L}_{\text{latent}}(\theta) + \lambda_2 \mathcal{L}_{\text{value}}(\theta) + \lambda_3 \mathcal{L}_{\text{end}}(\theta). 
\end{equation}
Guided by $\hat{A}_t$, this objective optimizes reasoning efficiency by rewarding shorter paths in predictable states while penalizing insufficient reasoning in scenarios requiring more cognitive planning.

\section{Experiments}

Following prior works~\cite{liu2025hybridvla,chen2025fast}, LaST-R1 is first pre-trained on a custom-designed large-scale dataset, detailed in Appendix \ref{appendix:pretrain}. We then systematically evaluate our framework: Section \ref{exp:se} compares our method against SOTA baselines on the LIBERO benchmark, and Section \ref{exp:ae} provides ablation studies on our core designs. Furthermore, Section \ref{exp:re} validates the performance of LaST-R1 in real-world robotic deployments, while Section \ref{exp:ge} thoroughly investigates its generalization capabilities.

\subsection{Simulation Experiment}
\label{exp:se}

\noindent \textbf{Settings.}
We validate our method on four LIBERO task suites \cite{liu2023libero}: LIBERO-Spatial, LIBERO-Object, LIBERO-Goal, and LIBERO-Long, each comprising 10 distinct tasks. All setups feature a simulated Franka Panda with a single front-view observation at 256 $\times$ 256 resolution. 
Following~\cite{li2025simplevla}, we adopt a two-stage training pipeline. 
We first perform an SFT warm-up using a single randomly selected data per task, followed by online RL to iteratively collect rollouts and update the policy within the simulator.
More training details are placed in Appendix \ref{appendix:warmup} and \ref{appendix:rl_libero}. Finally, performance is measured by the average Success Rate (SR) across 50 held-out test scenarios for each task.

\noindent \textbf{Baselines.}
We conduct a comprehensive comparison against both SFT-only and RL-trained models. 
All SFT baselines are trained on the full training dataset (50 trajectories per task).
Among the RL-trained baselines, GRAPE \cite{zhang2024grape} utilizes DPO \cite{rafailov2023direct}, VLA-RL \cite{lu2025vla} and $\pi_{\text{RL}}$ \cite{chen2025pirl} rely on PPO \cite{schulman2017proximal}, TGRPO \cite{chen2025tgrpofinetuningvisionlanguageactionmodel}, RLinf-GRPO \cite{zang2025rlinf}, and SimpleVLA-RL \cite{li2025simplevla} are optimized via GRPO \cite{shao2024deepseekmath}.
All RL-trained baselines follow their official warm-up strategies, as detailed in Table~\ref{tab:libero_comparison}.

\begin{table*}[t]
\centering
% \vspace{-0.1cm}
\caption{\textbf{Comparison on the LIBERO.} For RL, we use single-trajectory warm-up and single-view data. Specifically, $\dagger$ denotes full-trajectory warm-up, and $\ddagger$ indicates two-camera views training.
}

\resizebox{\textwidth}{!}{
\begin{tabular}{l c cc cc cc cc cc}
\toprule
\multirow{2}{*}{\textbf{Models}}
& \multirow{2}{*}{\textbf{Paradigm}}
& \multicolumn{2}{c}{\textbf{Spatial}} 
& \multicolumn{2}{c}{\textbf{Object}} 
& \multicolumn{2}{c}{\textbf{Goal}} 
& \multicolumn{2}{c}{\textbf{Long}} 
& \multicolumn{2}{c}{\textbf{Average}} \\
\cmidrule(lr){3-4} \cmidrule(lr){5-6} \cmidrule(lr){7-8} \cmidrule(lr){9-10} \cmidrule(lr){11-12}
& & SR$\uparrow$ & Rank$\downarrow$ & SR$\uparrow$ & Rank$\downarrow$ & SR$\uparrow$ & Rank$\downarrow$ & SR$\uparrow$ & Rank$\downarrow$ & SR$\uparrow$ & Rank$\downarrow$ \\

\midrule
OpenVLA \cite{kim2024openvla} & SFT & 84.7 & 12 & 88.4 & 12 & 79.2 & 12 & 53.7 & 12 & 76.5 & 12 \\
GR00T-N1 \cite{bjorck2025gr00t} & SFT & 94.4 & 8 & 97.6 & 8 & 93.0 & 8 & 90.6 & 7 & 93.9 & 8 \\
$\pi_0$ \cite{black2024pi_0} & SFT & 96.8 & 7 & 98.8 & 4 & 95.8 & 7 & 85.2 & 8 & 94.2 & 7 \\
$\pi_{0.5}$ \cite{intelligence2025pi_} & SFT & 98.8 & 4 & 98.2 & 7 & 98.0 & 5 & 92.4 & 5 & 96.9 & 5 \\
OpenVLA-OFT \cite{kim2025fine} & SFT & 97.6 & 6 & 98.4 & 6 & 97.9 & 6 & 94.5 & 2 & 97.1 & 4 \\

\midrule
GRAPE$^\dagger$ \cite{zhang2024grape} & RL & 88.5 & 11 & 92.1 & 10 & 83.1 & 9 & 57.2 & 11 & 80.2 & 11 \\
TGRPO$^\dagger$ \cite{chen2025tgrpofinetuningvisionlanguageactionmodel} & RL & 90.4 & 9 & 92.2 & 9 & 81.0 & 11 & 59.2 & 10 & 80.7 & 10 \\
VLA-RL$^\dagger$ \cite{lu2025vla} & RL & 90.2 & 10 & 91.8 & 11 & 82.2 & 10 & 59.8 & 9 & 81.0 & 9 \\
SimpleVLA-RL \cite{li2025simplevla} & RL & 98.2 & 5 & 98.7 & 5 & 98.8 & 3 & 91.7 & 6 & 96.9 & 5 \\
RLinf-GRPO \cite{zang2025rlinf} & RL & 98.9 & 3 & 99.7 & 3 & 98.3 & 4 & 93.6 & 4 & 97.6 & 3 \\
$\pi_\text{RL}$$^\ddagger$ \cite{chen2025pirl} & RL & 99.6 & 2 & \textbf{100.0} & 1 & 99.6 & 2 & 94.0 & 3 & 98.3 & 2 \\
\rowcolor{gray!15} LaST-R1 (Ours) & RL & \textbf{99.8} & 1 & \textbf{100.0} & 1 & \textbf{100.0} & 1 & \textbf{99.8} & 1 & \textbf{99.9} & 1 \\

\bottomrule
\end{tabular}
}
% \vspace{-0.1cm}
\label{tab:libero_comparison}
\end{table*}

\begin{figure*}[t]
    \centering
    \vspace{-0.2cm}
    \includegraphics[width=0.99\textwidth]{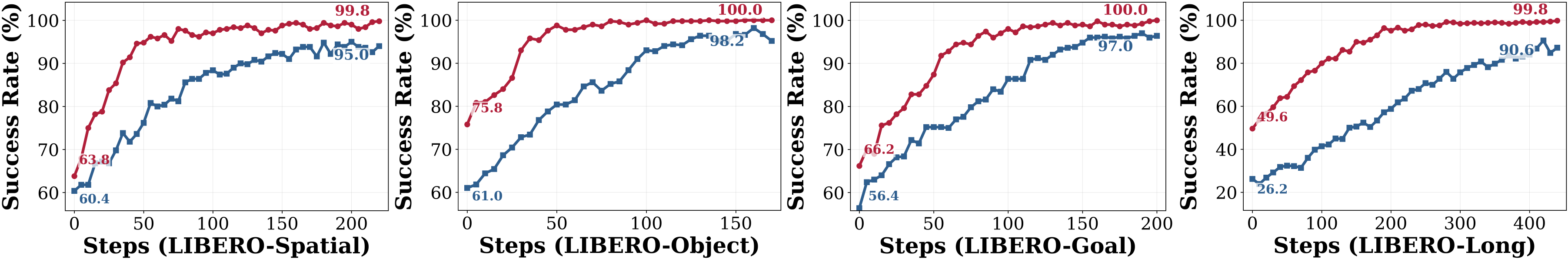} 
    \vspace{-0.25cm}
    \caption{\textbf{One-shot Online RL Learning Curves on LIBERO.}
    We compare our proposed LaST-R1 optimized via LAPO (\textcolor[HTML]{B11F3A}{red}), against the standard Action-Only baseline optimized via PPO (\textcolor[HTML]{2F5F8F}{blue}).}
    \label{fig:main_results}
    \vspace{-0.3cm}
\end{figure*}

\textbf{Result Analysis.}
Table \ref{tab:libero_comparison} presents the quantitative comparison across the LIBERO benchmark. Our proposed LaST-R1 achieves state-of-the-art performance, recording a near-perfect average success rate of 99.9\% and ranking first across all four task suites. Notably, despite using only a single trajectory for warm-up, our method outperforms strong SFT baselines such as $\pi_{0.5}$ (96.9\%) and OpenVLA-OFT (97.1\%), which heavily rely on complete expert datasets. When compared to the leading RL-based method $\pi_{\text{RL}}$ under the identical one-trajectory setting, LaST-R1 maintains a consistent advantage. This gap becomes particularly pronounced on the highly challenging LIBERO-Long suite (99.8\% vs. 94.0\%), demonstrating that our framework possesses superior capabilities in handling complex and long-horizon manipulation tasks.
Ultimately, these results demonstrate that LaST-R1 substantially improves the physical world modeling capability of VLA models, leading to near-perfect action performance even with one-shot training data.
Meanwhile, Figure~\ref{fig:main_results} presents the learning curves across four task suits in LIBERO. LaST-R1 trained with LAPO achieves significantly faster convergence and higher success rates compared to the Action-Only baseline optimized via PPO.
The baseline policy struggles with optimization efficiency, especially in the LIBERO-Long task suite, requiring significantly more steps yet still converging to sub-optimal performance.
The results show that introducing latent CoT optimization during post-training acts as an effective cognitive buffer, smoothing the RL optimization landscape and improving sample efficiency.
Furthermore, additional analyses are provided in Appendix \ref{appd:adaptive_length} and \ref{appd:episode_length}, showing how this optimized reasoning mechanism yields both adaptive reasoning lengths across tasks and optimized execution episode lengths.

\subsection{Ablation Studies}
\label{exp:ae}

\begin{figure*}[t]
    \centering
    \vspace{-0.1cm}
    \includegraphics[width=0.99\textwidth]{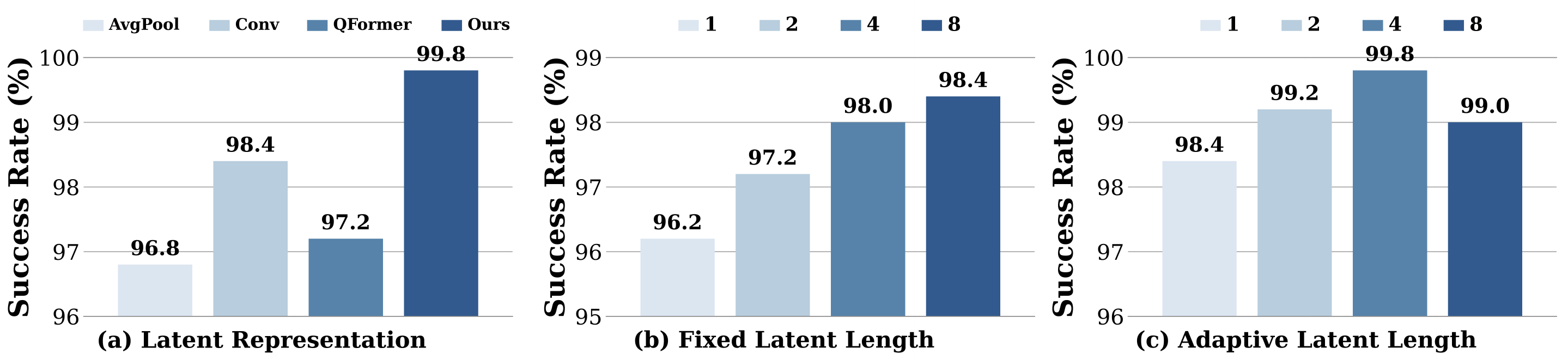} 
    \vspace{-0.3cm}
    \caption{\textbf{Ablation Studies.} We evaluate (a) latent representation methods, (b) different fixed latent CoT lengths, and (c) adaptive CoT length with varying \texttt{<latent\_end>} emission strategy.}
    \label{fig:ablation_latent}
    \vspace{-0.35cm}
\end{figure*}

To validate our core designs, we conduct ablation studies evaluated on the LIBERO-Spatial suite unless otherwise specified.
\noindent \textbf{(1) Effectiveness of Latent Reasoning in SFT and RL Post-Training.}
As shown in Figure~\ref{fig:main_results}, one-shot SFT with latent reasoning (LaST-R1 Warm-up) achieves a 63.9\% average SR, surpassing the Action-Only baseline (51.0\%). The improvement is particularly pronounced on LIBERO-Long (long-horizon tasks), where incorporating latent CoT boosts performance from 26.2\% to 49.6\%. After RL post-training, our full method (LaST-R1 + LAPO) further amplifies this advantage, achieving 99.9\% SR across all task suites and significantly outperforming Action-Only + PPO (95.2\%). These results demonstrate that our latent reasoning mechanism enhances physical understanding, leading to more effective action optimization, as further evidenced by action-to-vision attention (Appendix \ref{appd:a2v attn}).
\noindent \textbf{(2) Design Choices for Latent Representation.}
As shown in Figure \ref{fig:ablation_latent} (a), we study the impact of different latent representation construction methods by comparing our DINOv3-based approach with three alternatives: averaging VLA visual encoder features~\cite{zhai2023sigmoid} via global pooling~\cite{liu2026last}, downsampling them with convolution~\cite{cai2026internvla}, and extracting latents using a Q-Former~\cite{li2023blip}.
Detailed implementations of these variants are deferred to Appendix \ref{appendix:latent_detail}. Evaluated after RL post-training, our DINOv3 representation achieves the highest SR of 99.8\%, outperforming Convolution (98.4\%), Q-Former (97.2\%), and Global Pooling (96.8\%). 
This validates that leveraging pre-trained global representations from DINOv3 better preserves intricate spatial and semantic structures, while the top-$k$ selection scheme maximally retains dynamic information.
\noindent \textbf{(3) Impact of Latent Reasoning Length.}
We ablate the CoT horizon by varying the fixed latent sequence length $N_z$ $\in \{1, 2, 4, 8\}$, as shown in Figure \ref{fig:ablation_latent} (b). Compared to the Action-Only baseline (no latent reasoning, 95.0\% SR), performance monotonically increases from 96.2\% (1 token) to 98.4\% (8 tokens). 
This indicates that a longer cognitive horizon effectively encodes richer condition information for downstream actions.
Moreover, we observe that the performance gain from 4 to 8 tokens is marginal. To strike an optimal balance between task accuracy and inference speed, we cap the maximum reasoning sequence at 8 tokens.
\noindent \textbf{(4) Adaptive CoT Length and \texttt{<latent\_end>} Emission Strategy.}
Finally, we validate our adaptive latent CoT mechanism in Figure \ref{fig:ablation_latent} (c). Given a maximum reasoning length $N_{\text{max}} = 8$, we ablate the number of valid candidate positions $M$ for emitting \texttt{<latent\_end>}. Specifically, we evaluate $M \in \{1, 2, 4, 8\}$ by uniformly distributing the candidate positions across the latent sequence. For example, $M=1$ allows emission only after the 8th latent token, whereas $M=8$ permits emission after every latent token.
We find that $M=4$ achieves the best performance (99.8\%). While adaptive termination generally improves both efficiency and performance over the fixed-length baseline (98.4\%), setting $M$ too large ($M=8$, 99.0\%) slightly degrades performance.
Therefore, a moderately constrained adaptive strategy ($M=4$) offers the best trade-off between reasoning flexibility and optimization stability.
More ablation studies are provided in Appendix \ref{appendix:ablation}.

\subsection{Real-World Experiment}
\label{exp:re}

\textbf{Settings.}
We validate our method on four practical real-world manipulation tasks using Franka Research 3 arms, including one single-arm task and three dual-arm tasks: (1) \textit{Insert hexagon block}, (2) \textit{Open bag zipper}, (3) \textit{Wipe vase with sponge}, and (4) \textit{Open bottle cap}. For each task, we collect expert demonstrations using a space mouse.
Given the significant domain gap between simulation and real-world deployment, we follow recent real-world VLA RL methods~\cite{chen2025conrft,xu2026twinrl} and adopt our proposed \textit{latent reasoning-before-acting} RL framework.
To improve RL update efficiency, we incorporate Low-Rank Adaptation (LoRA)~\cite{hu2022lora} into all attention layers of LaST-R1, updating only the LoRA parameters.
Additional implementation details are provided in Appendix~\ref{ap:Real-World RL}.
For observations, we use one third-person camera and two wrist cameras, with input resolutions matching those in simulation. All policies are evaluated over 20 rollouts under varied positions, with task success determined by human evaluation.
To highlight the advantages of our RL post-training framework, we compare LaST-R1 with the SOTA VLA model $\pi_{0.5}$, trained via full SFT on 100 expert trajectories.

\begin{table*}[t]
\centering
\caption{
\textbf{Comparison on Real-World Tasks.} Evaluation includes the original training scenario and three generalization settings: unseen objects, background variations, and lighting conditions. Example cases are shown below, with changed configurations highlighted in \textcolor[HTML]{B11F3A}{red boxes}.
}
\small
\resizebox{\textwidth}{!}{
\begin{tabular}{lcccc|cccc}
\toprule
\multirow{2}{*}{\textbf{Methods}} & \multicolumn{4}{c|}{\textbf{Insert hexagon block}} & \multicolumn{4}{c}{\textbf{Open bag zipper}} \\
\cmidrule(lr){2-5} \cmidrule(lr){6-9}
& Original & Unseen-Object & -Background & -Lighting & Original & Unseen-Object & -Background & -Lighting \\
\midrule
$\pi_{0.5}$ \cite{intelligence2025pi_} (Full-size SFT) & 65 & 35 \textcolor{gray}{(-30\%)} &  55 \textcolor{gray}{(-10\%)} & 40 \textcolor{gray}{(-25\%)} & 75 & 30 \textcolor{gray}{(-45\%)} & 70 \textcolor{gray}{(-5\%)} & 60 \textcolor{gray}{(-15\%)} \\
LaST-R1 (Few-shot SFT $\rightarrow$ RL)    & 45 $\rightarrow$ 90 & 75 \textcolor{gray}{(-15\%)} & 85 \textcolor{gray}{(-5\%)} & 80 \textcolor{gray}{(-10\%)} & 55 $\rightarrow$ 95 & 80 \textcolor{gray}{(-15\%)} & 95 \textcolor{gray}{(-0\%)} & 90 \textcolor{gray}{(-5\%)}\\
\midrule
\multirow{2}{*}{\textbf{Methods}} & \multicolumn{4}{c|}{\textbf{Wipe vase with sponge}} & \multicolumn{4}{c}{\textbf{Open bottle cap}} \\
\cmidrule(lr){2-5} \cmidrule(lr){6-9}
& Original & Unseen-Object & -Background & -Lighting & Original & Unseen-Object & -Background & -Lighting \\
\midrule
$\pi_{0.5}$ \cite{intelligence2025pi_} (Full-size SFT) &  75 & 45 \textcolor{gray}{(-30\%)} & 65 \textcolor{gray}{(-10\%)} & 50 \textcolor{gray}{(-25\%)} & 70  & 50 \textcolor{gray}{(-20\%)} & 55 \textcolor{gray}{(-15\%)} & 55 \textcolor{gray}{(-15\%)}  \\
LaST-R1 (Few-shot SFT $\rightarrow$ RL)   & 65 $\rightarrow$ 95 & 80 \textcolor{gray}{(-15\%)} & 90 \textcolor{gray}{(-5\%)} & 95 \textcolor{gray}{(-0\%)} & 45 $\rightarrow$ 95 & 95 \textcolor{gray}{(-0\%)} & 80 \textcolor{gray}{(-15\%)} & 85 \textcolor{gray}{(-10\%)} \\
\bottomrule
\end{tabular}}
\includegraphics[width=\textwidth]{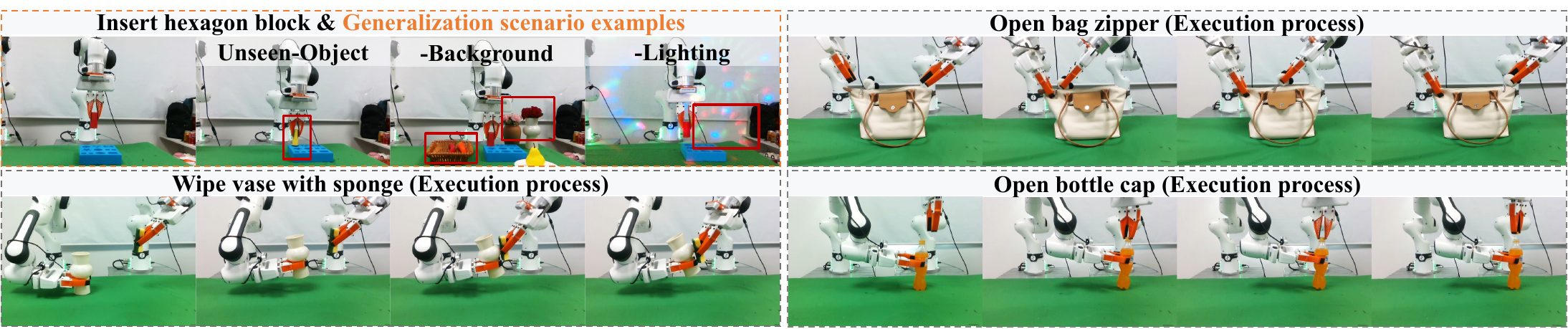}
\label{tab:realworld_results}
\vspace{-0.2cm}
\end{table*}

\textbf{Result Analysis.}
The “Original” column in Table \ref{tab:realworld_results} demonstrates the remarkable data efficiency of LaST-R1, where an initial warm-up with only 30 trajectories followed by online RL significantly outperforms the $\pi_{0.5}$ baseline trained with 100 expert trajectories via full supervised fine-tuning.
Overall, this paradigm boosts the average SR substantially from an initial 52.5\% to a final 93.75\% ($\pm$ 1.25\% std), surpassing the 71.25\% ($\pm$ 4.5\% std) average SR of the baseline $\pi_{0.5}$. 
The standard deviation (std) is computed over three independent runs.
Specifically, the proposed RL method enhances visually guided precision in single-arm tasks by doubling the success rate of \textit{Insert hexagon block} from 45\% to 90\%, while also improving dual-arm coordination for contact-rich manipulation like the \textit{Open bag zipper} and \textit{Wipe vase with sponge} tasks.
Furthermore, the dramatic performance surge from 45\% to 95\% in the \textit{Open bottle cap} task indicates that our post-training fundamentally strengthens the policy's ability to model temporally evolving physical interactions.
More visualizations and real-world execution videos are provided in Appendix \ref{appd:real_v} and the supplementary material.

\subsection{Generalization Analysis}
\label{exp:ge}

\begin{figure*}[t]
    \centering
    \vspace{-0.25cm}
    \includegraphics[width=\textwidth]{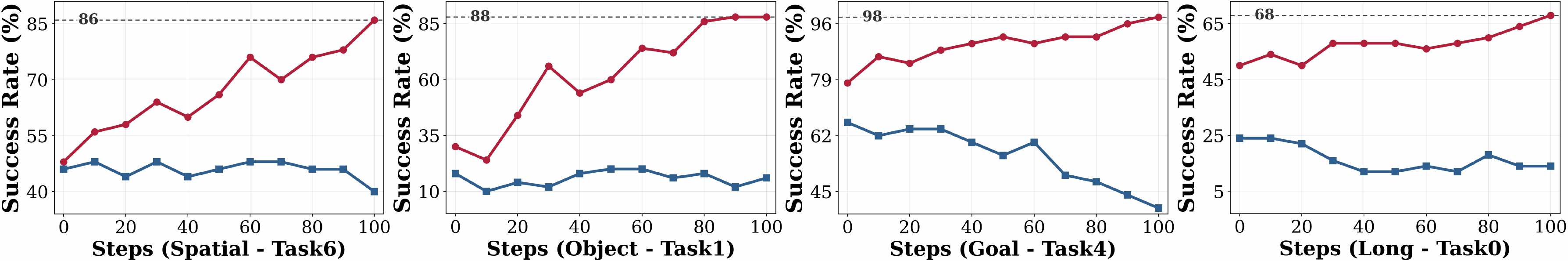} 
    \vspace{-0.6cm}
    \caption{
    \textbf{Generalization Analysis on LIBERO.} While the OOD performance of the Action-Only PPO baseline (\textcolor[HTML]{2F5F8F}{blue}) stagnates, our LaST-R1 with LAPO (\textcolor[HTML]{B11F3A}{red}) demonstrates continuous improvement.
    }
    \label{fig:ablation_gen}
    \vspace{-0.2cm}
\end{figure*}

We evaluated LaST-R1's generalization in both simulation and the real world.
\noindent \textbf{(1) Simulation.}
Following the setup of prior work~\cite{li2025simplevla}, for each task suite in LIBERO, we perform online RL on 9 “seen” tasks and hold out 1 task for out-of-distribution (OOD) evaluation, matching our one-trajectory warm-up setting. 
As shown in Figure \ref{fig:ablation_gen}, the standard Action-Only + PPO baseline suffers from severe overfitting, its OOD performance stagnates and even degrades on the LIBERO-Goal and LIBERO-Long suites. Conversely, our LaST-R1 + LAPO policy demonstrates continuous, significant OOD improvements across all suites. 
This demonstrates that our latent reasoning with LAPO progressively improve the model’s understanding of scene and physical dynamics via environmental interaction, enabling it to extract transferable spatial and semantic concepts rather than overfitting to the training distribution.
Additional experiments are presented in Appendix~\ref{ap:AGA}.
\noindent \textbf{(2) Real-World.}
This superior adaptability seamlessly transfers to physical deployments. As shown in Table \ref{tab:realworld_results}, when exposed to severe unseen perturbations, the full-size SFT baseline $\pi_{0.5}$ suffers catastrophic performance degradation, evidenced by up to a 45\% drop when interacting with unseen objects. In stark contrast, LaST-R1 (Few-shot SFT $\rightarrow$ RL) maintains highly resilient execution. It successfully confines performance drops to within 15\% for novel objects and demonstrates near-zero degradation against diverse background and lighting variations. This confirms that optimizing latent reasoning through environmental feedback intrinsically fortifies LaST-R1 VLA against environmental uncertainties.

%%%%%%%%%%%%%%%%%%%%%%%%%%%%%%%%%%%%%%%%%%%%%%%%%%%%%%%%%%%%

\section{Conclusion}
In this work, we present LaST-R1, a tailored RL post-training framework for ``latent reasoning-before-acting'' policies with a unified VLA model coupling latent reasoning and action generation. By introducing latent Chain-of-Thought reasoning grounded in global visual representations, our approach enables structured modeling of physical dynamics prior to action execution. Building on this design, we propose Latent-to-Action Policy Optimization (LAPO), a novel RL paradigm that jointly optimizes reasoning and action spaces, allowing environmental feedback to shape both external behavior and internal reasoning processes. Furthermore, we introduce an adaptive latent CoT mechanism that dynamically adjusts the reasoning horizon based on task diversity, improving both efficiency and performance.
Experimental results highlight the importance of integrating physical latent reasoning into VLA learning and show that jointly optimizing reasoning and action through RL is a promising direction for improving generalization and physical understanding. 
We hope this work provides a foundation for scaling latent reasoning-driven policies in complex robotic manipulation.

%%%%%%%%%%%%%%%%%%%%%%%%%%%%%%%%%%%%%%%%%%%%%%%%%%%%%%%%%%%%

\bibliographystyle{unsrt}
\bibliography{ref}

@article{schulman2017proximal,
  title={Proximal policy optimization algorithms},
  author={Schulman, John and Wolski, Filip and Dhariwal, Prafulla and Radford, Alec and Klimov, Oleg},
  journal={arXiv preprint arXiv:1707.06347},
  year={2017}
}

@article{wu2024robomind,
  title={Robomind: Benchmark on multi-embodiment intelligence normative data for robot manipulation},
  author={Wu, Kun and Hou, Chengkai and Liu, Jiaming and Che, Zhengping and Ju, Xiaozhu and Yang, Zhuqin and Li, Meng and Zhao, Yinuo and Xu, Zhiyuan and Yang, Guang and others},
  journal={arXiv preprint arXiv:2412.13877},
  year={2024}
}

@article{tschannen2025siglip,
  title={Siglip 2: Multilingual vision-language encoders with improved semantic understanding, localization, and dense features},
  author={Tschannen, Michael and Gritsenko, Alexey and Wang, Xiao and Naeem, Muhammad Ferjad and Alabdulmohsin, Ibrahim and Parthasarathy, Nikhil and Evans, Talfan and Beyer, Lucas and Xia, Ye and Mustafa, Basil and others},
  journal={arXiv preprint arXiv:2502.14786},
  year={2025}
}

@inproceedings{jang2022bc,
  title={Bc-z: Zero-shot task generalization with robotic imitation learning},
  author={Jang, Eric and Irpan, Alex and Khansari, Mohi and Kappler, Daniel and Ebert, Frederik and Lynch, Corey and Levine, Sergey and Finn, Chelsea},
  booktitle={Conference on Robot Learning},
  pages={991--1002},
  year={2022},
  organization={PMLR}
}

@misc{BerkeleyUR5Website,
  title = {Berkeley {UR5} Demonstration Dataset},
  author = {Lawrence Yunliang Chen and Simeon Adebola and Ken Goldberg},
  howpublished = {\url{https://sites.google.com/view/berkeley-ur5/home}},
}

@inproceedings{ebert2022bridge,
  title={Bridge data: Boosting generalization of robotic skills with cross-domain datasets},
  author={Ebert, Frederik and Yang, Yanlai and Schmeckpeper, Karl and Bucher, Bernadette and Georgakis, Georgios and Daniilidis, Kostas and Finn, Chelsea and Levine, Sergey},
  booktitle={RSS},
  year={2022}
}

@misc{walke2023bridgedata,
      title={BridgeData V2: A Dataset for Robot Learning at Scale}, 
      author={Homer Walke and Kevin Black and Abraham Lee and Moo Jin Kim and Max Du and Chongyi Zheng and Tony Zhao and Philippe Hansen-Estruch and Quan Vuong and Andre He and Vivek Myers and Kuan Fang and Chelsea Finn and Sergey Levine},
      year={2023},
      eprint={2308.12952},
      archivePrefix={arXiv},
      primaryClass={cs.RO}
}

@article{padalkar2023guided,
  title={A guided reinforcement learning approach using shared control templates for learning manipulation skills in the real world},
  author={Padalkar, Abhishek and Quere, Gabriel and Raffin, Antonin and Silv{\'e}rio, Jo{\~a}o and Stulp, Freek},
  year={2023}
}

@misc{shafiullah2023dobbe,
    title={On Bringing Robots Home}, 
    author={Nur Muhammad Mahi Shafiullah and Anant Rai and Haritheja Etukuru and Yiqian Liu and Ishan Misra and Soumith Chintala and Lerrel Pinto},
    year={2023},
    eprint={2311.16098},
    archivePrefix={arXiv},
    primaryClass={cs.RO}
}

@misc{luo2023fmb,
  title = {{FMB}: A Functional Manipulation Benchmark for Generalizable Robotic Learning},
  author={Jianlan Luo and Charles Xu and Fangchen Liu and Liam Tan and Zipeng Lin and Jeffrey Wu and Pieter Abbeel and Sergey Levine},
  year={2023},
  howpublished = {\url{https://functional-manipulation-benchmark.github.io}},
}

@misc{dass2023jacoplay,
  author = {Dass, Shivin and Yapeter, Jullian and Zhang, Jesse and Zhang, Jiahui
            and Pertsch, Karl and Nikolaidis, Stefanos and Lim, Joseph J.},
  title = {{CLVR} Jaco Play Dataset},
  url = {https://github.com/clvrai/clvr_jaco_play_dataset},
  version = {1.0.0},
  year = {2023}
}

@article{kalashnikov2018qt,
  title={{QT}-{O}pt: Scalable deep reinforcement learning for vision-based robotic manipulation},
  author={Kalashnikov, Dmitry and Irpan, Alex and Pastor, Peter and Ibarz, Julian and Herzog, Alexander and Jang, Eric and Quillen, Deirdre and Holly, Ethan and Kalakrishnan, Mrinal and Vanhoucke, Vincent and others},
  journal={arXiv preprint arXiv:1806.10293},
  year={2018}
}

@article{lynch2023interactive,
  title={Interactive language: Talking to robots in real time},
  author={Lynch, Corey and Wahid, Ayzaan and Tompson, Jonathan and Ding, Tianli and Betker, James and Baruch, Robert and Armstrong, Travis and Florence, Pete},
  journal={IEEE Robotics and Automation Letters},
  year={2023},
  publisher={IEEE}
}

@misc{gu2023maniskill2unifiedbenchmarkgeneralizable,
      title={ManiSkill2: A Unified Benchmark for Generalizable Manipulation Skills}, 
      author={Jiayuan Gu and Fanbo Xiang and Xuanlin Li and Zhan Ling and Xiqiang Liu and Tongzhou Mu and Yihe Tang and Stone Tao and Xinyue Wei and Yunchao Yao and Xiaodi Yuan and Pengwei Xie and Zhiao Huang and Rui Chen and Hao Su},
      year={2023},
      eprint={2302.04659},
      archivePrefix={arXiv},
      primaryClass={cs.RO},
      url={https://arxiv.org/abs/2302.04659}, 
}

@article{cui2022play,
  title   = {From Play to Policy: Conditional Behavior Generation from Uncurated Robot Data},
  author  = {Cui, Zichen Jeff and Wang, Yibin and Shafiullah, Nur Muhammad Mahi and Pinto, Lerrel},
  journal = {arXiv preprint arXiv:2210.10047},
  year    = {2022}
}

@inproceedings{dasari2020robonet,
  title={RoboNet: Large-Scale Multi-Robot Learning},
  author={Dasari, Sudeep and Ebert, Frederik and Tian, Stephen and Nair, Suraj and Bucher, Bernadette and Schmeckpeper, Karl and Singh, Siddharth and Levine, Sergey and Finn, Chelsea},
  booktitle={Conference on Robot Learning},
  pages={885--897},
  year={2020},
  organization={PMLR}
}

@inproceedings{
kumar2023robohive,
title={RoboHive: A Unified Framework for Robot Learning},
author={Vikash Kumar and Rutav Shah and Gaoyue Zhou and Vincent Moens and Vittorio Caggiano and Abhishek Gupta and Aravind Rajeswaran},
booktitle={Thirty-seventh Conference on Neural Information Processing Systems Datasets and Benchmarks Track},
year={2023}
}

@article{DBLP:journals/corr/abs-1811-02790,
  author       = {Ajay Mandlekar and
                  Yuke Zhu and
                  Animesh Garg and
                  Jonathan Booher and
                  Max Spero and
                  Albert Tung and
                  Julian Gao and
                  John Emmons and
                  Anchit Gupta and
                  Emre Orbay and
                  Silvio Savarese and
                  Li Fei{-}Fei},
  title        = {{R}obo{T}urk: {A} Crowdsourcing Platform for Robotic Skill Learning through
                  Imitation},
  journal      = {CoRR},
  volume       = {abs/1811.02790},
  year         = {2018},
  eprinttype    = {arXiv},
  eprint       = {1811.02790},
  bibsource    = {dblp computer science bibliography, https://dblp.org}
}

@article{belkhale2023hydra,
 title={HYDRA: Hybrid Robot Actions for Imitation Learning},
 author={Belkhale, Suneel and Cui, Yuchen and Sadigh, Dorsa},
 journal={arxiv},
 year={2023}
}

@inproceedings{rosetebeas2022latent,
author = {Erick Rosete-Beas and Oier Mees and Gabriel Kalweit and Joschka Boedecker and Wolfram Burgard},
title = {Latent Plans for Task Agnostic Offline Reinforcement Learning},
booktitle = {Proceedings of the 6th Conference on Robot Learning (CoRL)},
year = {2022}
}

@inproceedings{mees2023grounding,
  title={Grounding  Language  with  Visual  Affordances  over  Unstructured  Data},
  author={Oier Mees and Jessica Borja-Diaz and Wolfram Burgard},
  booktitle = {Proceedings of the IEEE International Conference on Robotics and Automation (ICRA)},
  year={2023},
  address = {London, UK}
}

@inproceedings{luo2024serl,
  title={Serl: A software suite for sample-efficient robotic reinforcement learning},
  author={Luo, Jianlan and Hu, Zheyuan and Xu, Charles and Tan, You Liang and Berg, Jacob and Sharma, Archit and Schaal, Stefan and Finn, Chelsea and Gupta, Abhishek and Levine, Sergey},
  booktitle={2024 IEEE International Conference on Robotics and Automation (ICRA)},
  pages={16961--16969},
  year={2024},
  organization={IEEE}
}

@article{luo2025precise,
  title={Precise and dexterous robotic manipulation via human-in-the-loop reinforcement learning},
  author={Luo, Jianlan and Xu, Charles and Wu, Jeffrey and Levine, Sergey},
  journal={Science Robotics},
  volume={10},
  number={105},
  pages={eads5033},
  year={2025},
  publisher={American Association for the Advancement of Science}
}

@article{luo2026look,
  title={Look Before Acting: Enhancing Vision Foundation Representations for Vision-Language-Action Models},
  author={Luo, Yulin and Chen, Hao and Wu, Zhuangzhe and Sui, Bowen and Liu, Jiaming and Gu, Chenyang and Liu, Zhuoyang and Feng, Qiuxuan and Yu, Jiale and Gu, Shuo and others},
  journal={arXiv preprint arXiv:2603.15618},
  year={2026}
}

@inproceedings{selvaraju2017grad,
  title={Grad-cam: Visual explanations from deep networks via gradient-based localization},
  author={Selvaraju, Ramprasaath R and Cogswell, Michael and Das, Abhishek and Vedantam, Ramakrishna and Parikh, Devi and Batra, Dhruv},
  booktitle={Proceedings of the IEEE international conference on computer vision},
  pages={618--626},
  year={2017}
}

@article{loshchilov2017decoupled,
  title={Decoupled weight decay regularization},
  author={Loshchilov, Ilya and Hutter, Frank},
  journal={arXiv preprint arXiv:1711.05101},
  year={2017}
}

@Misc{accelerate,
  title =        {Accelerate: Training and inference at scale made simple, efficient and adaptable.},
  author =       {Sylvain Gugger and Lysandre Debut and Thomas Wolf and Philipp Schmid and Zachary Mueller and Sourab Mangrulkar and Marc Sun and Benjamin Bossan},
  howpublished = {\url{https://github.com/huggingface/accelerate}},
  year =         {2022}
}

@misc{zhou2023train,
      title={Train Offline, Test Online: A Real Robot Learning Benchmark}, 
      author={Gaoyue Zhou and Victoria Dean and Mohan Kumar Srirama and Aravind Rajeswaran and Jyothish Pari and Kyle Hatch and Aryan Jain and Tianhe Yu and Pieter Abbeel and Lerrel Pinto and Chelsea Finn and Abhinav Gupta},
      year={2023},
      eprint={2306.00942},
      archivePrefix={arXiv},
      primaryClass={cs.RO}
}

@misc{oh2023pr2utokyodatasets,
  author={Jihoon Oh and Naoaki Kanazawa and Kento Kawaharazuka},
  title={X-Embodiment U-Tokyo PR2 Datasets},
  year={2023},
  url={https://github.com/ojh6404/rlds_dataset_builder},
}

@misc{matsushima2023weblab,
  title={Weblab xArm Dataset},
  author={Tatsuya Matsushima and Hiroki Furuta and Yusuke Iwasawa and Yutaka Matsuo},
  year={2023},
}

@inproceedings{rt12022arxiv,
    title={RT-1: Robotics Transformer for Real-World Control at Scale},
    author={Anthony	Brohan and  Noah Brown and  Justice Carbajal and  Yevgen Chebotar and  Joseph Dabis and  Chelsea Finn and  Keerthana Gopalakrishnan and  Karol Hausman and  Alex Herzog and  Jasmine Hsu and  Julian Ibarz and others},
    booktitle={arXiv preprint arXiv:2212.06817},
    year={2022}
}

@inproceedings{heo2023furniturebench,
  title={FurnitureBench: Reproducible Real-World Benchmark for Long-Horizon Complex Manipulation},
  author={Minho Heo and Youngwoon Lee and Doohyun Lee and Joseph J. Lim},
  booktitle={Robotics: Science and Systems},
  year={2023}
}

@article{mendonca2023structured,
  title={Structured World Models from Human Videos},
  author={Mendonca, Russell and Bahl, Shikhar and Pathak, Deepak},
  journal={CoRL},
  year={2023}
}

@inproceedings{sheng2025hybridflow,
  title={Hybridflow: A flexible and efficient rlhf framework},
  author={Sheng, Guangming and Zhang, Chi and Ye, Zilingfeng and Wu, Xibin and Zhang, Wang and Zhang, Ru and Peng, Yanghua and Lin, Haibin and Wu, Chuan},
  booktitle={Proceedings of the Twentieth European Conference on Computer Systems},
  pages={1279--1297},
  year={2025}
}

@article{khazatsky2024droid,
  title={Droid: A large-scale in-the-wild robot manipulation dataset},
  author={Khazatsky, Alexander and Pertsch, Karl and Nair, Suraj and Balakrishna, Ashwin and Dasari, Sudeep and Karamcheti, Siddharth and Nasiriany, Soroush and Srirama, Mohan Kumar and Chen, Lawrence Yunliang and Ellis, Kirsty and others},
  journal={arXiv preprint arXiv:2403.12945},
  year={2024}
}

@inproceedings{o2024open,
  title={Open x-embodiment: Robotic learning datasets and rt-x models: Open x-embodiment collaboration 0},
  author={O’Neill, Abby and Rehman, Abdul and Maddukuri, Abhiram and Gupta, Abhishek and Padalkar, Abhishek and Lee, Abraham and Pooley, Acorn and Gupta, Agrim and Mandlekar, Ajay and Jain, Ajinkya and others},
  booktitle={2024 IEEE International Conference on Robotics and Automation (ICRA)},
  pages={6892--6903},
  year={2024},
  organization={IEEE}
}

@inproceedings{moritz2018ray,
  title={Ray: A distributed framework for emerging $\{$AI$\}$ applications},
  author={Moritz, Philipp and Nishihara, Robert and Wang, Stephanie and Tumanov, Alexey and Liaw, Richard and Liang, Eric and Elibol, Melih and Yang, Zongheng and Paul, William and Jordan, Michael I and others},
  booktitle={13th USENIX symposium on operating systems design and implementation (OSDI 18)},
  pages={561--577},
  year={2018}
}

@inproceedings{li2023blip,
  title={Blip-2: Bootstrapping language-image pre-training with frozen image encoders and large language models},
  author={Li, Junnan and Li, Dongxu and Savarese, Silvio and Hoi, Steven},
  booktitle={International conference on machine learning},
  pages={19730--19742},
  year={2023},
  organization={PMLR}
}

@article{hu2022lora,
  title={Lora: Low-rank adaptation of large language models.},
  author={Hu, Edward J and Shen, Yelong and Wallis, Phillip and Allen-Zhu, Zeyuan and Li, Yuanzhi and Wang, Shean and Wang, Liang and Chen, Weizhu and others},
  journal={Iclr},
  volume={1},
  number={2},
  pages={3},
  year={2022}
}

@article{liu2023libero,
  title={Libero: Benchmarking knowledge transfer for lifelong robot learning},
  author={Liu, Bo and Zhu, Yifeng and Gao, Chongkai and Feng, Yihao and Liu, Qiang and Zhu, Yuke and Stone, Peter},
  journal={Advances in Neural Information Processing Systems},
  volume={36},
  pages={44776--44791},
  year={2023}
}

@article{bai2026latent,
  title={Latent Reasoning VLA: Latent Thinking and Prediction for Vision-Language-Action Models},
  author={Bai, Shuanghao and Lyu, Jing and Zhou, Wanqi and Li, Zhe and Wang, Dakai and Xing, Lei and Zhao, Xiaoguang and Wang, Pengwei and Wang, Zhongyuan and Chi, Cheng and others},
  journal={arXiv preprint arXiv:2602.01166},
  year={2026}
}

@article{chen2025pirl,
  title={$\pi$RL: Online rl fine-tuning for flow-based vision-language-action models},
  author={Chen, Kang and Liu, Zhihao and Zhang, Tonghe and Guo, Zhen and Xu, Si and Lin, Hao and Zang, Hongzhi and Zhang, Quanlu and Yu, Zhaofei and Fan, Guoliang and others},
  journal={arXiv preprint arXiv:2510.25889},
  year={2025}
}

@article{liu2025can,
  title={What can rl bring to vla generalization? an empirical study},
  author={Liu, Jijia and Gao, Feng and Wei, Bingwen and Chen, Xinlei and Liao, Qingmin and Wu, Yi and Yu, Chao and Wang, Yu},
  journal={arXiv preprint arXiv:2505.19789},
  year={2025}
}

@article{shao2024deepseekmath,
  title={Deepseekmath: Pushing the limits of mathematical reasoning in open language models},
  author={Shao, Zhihong and Wang, Peiyi and Zhu, Qihao and Xu, Runxin and Song, Junxiao and Bi, Xiao and Zhang, Haowei and Zhang, Mingchuan and Li, YK and Wu, Yang and others},
  journal={arXiv preprint arXiv:2402.03300},
  year={2024}
}

@article{lu2025vla,
  title={Vla-rl: Towards masterful and general robotic manipulation with scalable reinforcement learning},
  author={Lu, Guanxing and Guo, Wenkai and Zhang, Chubin and Zhou, Yuheng and Jiang, Haonan and Gao, Zifeng and Tang, Yansong and Wang, Ziwei},
  journal={arXiv preprint arXiv:2505.18719},
  year={2025}
}

@article{li2025simplevla,
  title={Simplevla-rl: Scaling vla training via reinforcement learning},
  author={Li, Haozhan and Zuo, Yuxin and Yu, Jiale and Zhang, Yuhao and Yang, Zhaohui and Zhang, Kaiyan and Zhu, Xuekai and Zhang, Yuchen and Chen, Tianxing and Cui, Ganqu and others},
  journal={arXiv preprint arXiv:2509.09674},
  year={2025}
}

@article{rafailov2023direct,
  title={Direct preference optimization: Your language model is secretly a reward model},
  author={Rafailov, Rafael and Sharma, Archit and Mitchell, Eric and Manning, Christopher D and Ermon, Stefano and Finn, Chelsea},
  journal={Advances in neural information processing systems},
  volume={36},
  pages={53728--53741},
  year={2023}
}

@article{zhang2024grape,
  title={Grape: Generalizing robot policy via preference alignment},
  author={Zhang, Zijian and Zheng, Kaiyuan and Chen, Zhaorun and Jang, Joel and Li, Yi and Han, Siwei and Wang, Chaoqi and Ding, Mingyu and Fox, Dieter and Yao, Huaxiu},
  journal={arXiv preprint arXiv:2411.19309},
  year={2024}
}

@article{cai2026internvla,
  title={InternVLA-A1: Unifying Understanding, Generation and Action for Robotic Manipulation},
  author={Cai, Junhao and Cai, Zetao and Cao, Jiafei and Chen, Yilun and He, Zeyu and Jiang, Lei and Li, Hang and Li, Hengjie and Li, Yang and Liu, Yufei and others},
  journal={arXiv preprint arXiv:2601.02456},
  year={2026}
}

@article{ye2025vla,
  title={Vla-r1: Enhancing reasoning in vision-language-action models},
  author={Ye, Angen and Zhang, Zeyu and Wang, Boyuan and Wang, Xiaofeng and Zhang, Dapeng and Zhu, Zheng},
  journal={arXiv preprint arXiv:2510.01623},
  year={2025}
}

@article{huang2025thinkact,
  title={Thinkact: Vision-language-action reasoning via reinforced visual latent planning},
  author={Huang, Chi-Pin and Wu, Yueh-Hua and Chen, Min-Hung and Wang, Yu-Chiang Frank and Yang, Fu-En},
  journal={arXiv preprint arXiv:2507.16815},
  year={2025}
}

@article{wei2022chain,
  title={Chain-of-thought prompting elicits reasoning in large language models},
  author={Wei, Jason and Wang, Xuezhi and Schuurmans, Dale and Bosma, Maarten and Xia, Fei and Chi, Ed and Le, Quoc V and Zhou, Denny and others},
  journal={Advances in neural information processing systems},
  volume={35},
  pages={24824--24837},
  year={2022}
}

@article{kim2024openvla,
  title={Openvla: An open-source vision-language-action model},
  author={Kim, Moo Jin and Pertsch, Karl and Karamcheti, Siddharth and Xiao, Ted and Balakrishna, Ashwin and Nair, Suraj and Rafailov, Rafael and Foster, Ethan and Lam, Grace and Sanketi, Pannag and others},
  journal={arXiv preprint arXiv:2406.09246},
  year={2024}
}

@article{beyer2024paligemma,
  title={Paligemma: A versatile 3b vlm for transfer},
  author={Beyer, Lucas and Steiner, Andreas and Pinto, Andr{\'e} Susano and Kolesnikov, Alexander and Wang, Xiao and Salz, Daniel and Neumann, Maxim and Alabdulmohsin, Ibrahim and Tschannen, Michael and Bugliarello, Emanuele and others},
  journal={arXiv preprint arXiv:2407.07726},
  year={2024}
}

@article{kim2025fine,
  title={Fine-tuning vision-language-action models: Optimizing speed and success},
  author={Kim, Moo Jin and Finn, Chelsea and Liang, Percy},
  journal={arXiv preprint arXiv:2502.19645},
  year={2025}
}

@article{black2024pi_0,
  title={$\pi\_0$: A Vision-Language-Action Flow Model for General Robot Control},
  author={Black, Kevin and Brown, Noah and Driess, Danny and Esmail, Adnan and Equi, Michael and Finn, Chelsea and Fusai, Niccolo and Groom, Lachy and Hausman, Karol and Ichter, Brian and others},
  journal={arXiv preprint arXiv:2410.24164},
  year={2024}
}

@article{intelligence2025pi_,
  title={$\pi$\_$0.5$: a Vision-Language-Action Model with Open-World Generalization},
  author={Intelligence, Physical and Black, Kevin and Brown, Noah and Darpinian, James and Dhabalia, Karan and Driess, Danny and Esmail, Adnan and Equi, Michael and Finn, Chelsea and Fusai, Niccolo and others},
  journal={arXiv preprint arXiv:2504.16054},
  year={2025}
}

@article{intelligence2025pi,
  title={${\pi}_{0.6}^{*}$: a VLA That Learns From Experience},
  author={Intelligence, Physical and Amin, Ali and Aniceto, Raichelle and Balakrishna, Ashwin and Black, Kevin and Conley, Ken and Connors, Grace and Darpinian, James and Dhabalia, Karan and DiCarlo, Jared and others},
  journal={arXiv preprint arXiv:2511.14759},
  year={2025}
}

@article{liu2025hybridvla,
  title={Hybridvla: Collaborative diffusion and autoregression in a unified vision-language-action model},
  author={Liu, Jiaming and Chen, Hao and An, Pengju and Liu, Zhuoyang and Zhang, Renrui and Gu, Chenyang and Li, Xiaoqi and Guo, Ziyu and Chen, Sixiang and Liu, Mengzhen and others},
  journal={arXiv preprint arXiv:2503.10631},
  year={2025}
}

@article{chen2025fast,
  title={Fast-in-slow: A dual-system foundation model unifying fast manipulation within slow reasoning},
  author={Chen, Hao and Liu, Jiaming and Gu, Chenyang and Liu, Zhuoyang and Zhang, Renrui and Li, Xiaoqi and He, Xiao and Guo, Yandong and Fu, Chi-Wing and Zhang, Shanghang and others},
  journal={arXiv preprint arXiv:2506.01953},
  year={2025}
}

@article{liu2026last,
  title={LaST $ \_ $\{$0$\}$ $: Latent Spatio-Temporal Chain-of-Thought for Robotic Vision-Language-Action Model},
  author={Liu, Zhuoyang and Liu, Jiaming and Chen, Hao and Yu, Jiale and Guo, Ziyu and Hou, Chengkai and Gu, Chenyang and Mi, Xiangju and Zhang, Renrui and Wu, Kun and others},
  journal={arXiv preprint arXiv:2601.05248},
  year={2026}
}

@article{li2024cogact,
  title={Cogact: A foundational vision-language-action model for synergizing cognition and action in robotic manipulation},
  author={Li, Qixiu and Liang, Yaobo and Wang, Zeyu and Luo, Lin and Chen, Xi and Liao, Mozheng and Wei, Fangyun and Deng, Yu and Xu, Sicheng and Zhang, Yizhong and others},
  journal={arXiv preprint arXiv:2411.19650},
  year={2024}
}

@article{karamcheti2024prismatic,
  title={Prismatic vlms: Investigating the design space of visually-conditioned language models},
  author={Karamcheti, Siddharth and Nair, Suraj and Balakrishna, Ashwin and Liang, Percy and Kollar, Thomas and Sadigh, Dorsa},
  journal={arXiv preprint arXiv:2402.07865},
  year={2024}
}

@misc{bai2025qwen3vltechnicalreport,
      title={Qwen3-VL Technical Report}, 
      author={Shuai Bai and Yuxuan Cai and Ruizhe Chen and Keqin Chen and Xionghui Chen and Zesen Cheng and Lianghao Deng and Wei Ding and Chang Gao and Chunjiang Ge and Wenbin Ge and Zhifang Guo and Qidong Huang and Jie Huang and Fei Huang and Binyuan Hui and Shutong Jiang and Zhaohai Li and Mingsheng Li and Mei Li and Kaixin Li and Zicheng Lin and Junyang Lin and Xuejing Liu and Jiawei Liu and Chenglong Liu and Yang Liu and Dayiheng Liu and Shixuan Liu and Dunjie Lu and Ruilin Luo and Chenxu Lv and Rui Men and Lingchen Meng and Xuancheng Ren and Xingzhang Ren and Sibo Song and Yuchong Sun and Jun Tang and Jianhong Tu and Jianqiang Wan and Peng Wang and Pengfei Wang and Qiuyue Wang and Yuxuan Wang and Tianbao Xie and Yiheng Xu and Haiyang Xu and Jin Xu and Zhibo Yang and Mingkun Yang and Jianxin Yang and An Yang and Bowen Yu and Fei Zhang and Hang Zhang and Xi Zhang and Bo Zheng and Humen Zhong and Jingren Zhou and Fan Zhou and Jing Zhou and Yuanzhi Zhu and Ke Zhu},
      year={2025},
      eprint={2511.21631},
      archivePrefix={arXiv},
      primaryClass={cs.CV},
      url={https://arxiv.org/abs/2511.21631}, 
}

@article{simeoni2025dinov3,
  title={Dinov3},
  author={Sim{\'e}oni, Oriane and Vo, Huy V and Seitzer, Maximilian and Baldassarre, Federico and Oquab, Maxime and Jose, Cijo and Khalidov, Vasil and Szafraniec, Marc and Yi, Seungeun and Ramamonjisoa, Micha{\"e}l and others},
  journal={arXiv preprint arXiv:2508.10104},
  year={2025}
}

@inproceedings{zhao2025cot,
  title={Cot-vla: Visual chain-of-thought reasoning for vision-language-action models},
  author={Zhao, Qingqing and Lu, Yao and Kim, Moo Jin and Fu, Zipeng and Zhang, Zhuoyang and Wu, Yecheng and Li, Zhaoshuo and Ma, Qianli and Han, Song and Finn, Chelsea and others},
  booktitle={Proceedings of the Computer Vision and Pattern Recognition Conference},
  pages={1702--1713},
  year={2025}
}

@article{qu2025spatialvla,
  title={Spatialvla: Exploring spatial representations for visual-language-action model},
  author={Qu, Delin and Song, Haoming and Chen, Qizhi and Yao, Yuanqi and Ye, Xinyi and Ding, Yan and Wang, Zhigang and Gu, JiaYuan and Zhao, Bin and Wang, Dong and others},
  journal={arXiv preprint arXiv:2501.15830},
  year={2025}
}

@article{bjorck2025gr00t,
  title={Gr00t n1: An open foundation model for generalist humanoid robots},
  author={Bjorck, Johan and Casta{\~n}eda, Fernando and Cherniadev, Nikita and Da, Xingye and Ding, Runyu and Fan, Linxi and Fang, Yu and Fox, Dieter and Hu, Fengyuan and Huang, Spencer and others},
  journal={arXiv preprint arXiv:2503.14734},
  year={2025}
}

@article{cen2025worldvla,
  title={Worldvla: Towards autoregressive action world model},
  author={Cen, Jun and Yu, Chaohui and Yuan, Hangjie and Jiang, Yuming and Huang, Siteng and Guo, Jiayan and Li, Xin and Song, Yibing and Luo, Hao and Wang, Fan and others},
  journal={arXiv preprint arXiv:2506.21539},
  year={2025}
}

@article{pan2026sop,
  title={SOP: A Scalable Online Post-Training System for Vision-Language-Action Models},
  author={Pan, Mingjie and Feng, Siyuan and Zhang, Qinglin and Li, Xinchen and Song, Jianheng and Qu, Chendi and Wang, Yi and Li, Chuankang and Xiong, Ziyu and Chen, Zhi and others},
  journal={arXiv preprint arXiv:2601.03044},
  year={2026}
}

@article{xu2026twinrl,
  title={TwinRL-VLA: Digital Twin-Driven Reinforcement Learning for Real-World Robotic Manipulation},
  author={Xu, Qinwen and Liu, Jiaming and Zhou, Rui and Shi, Shaojun and Han, Nuowei and Liu, Zhuoyang and Gu, Chenyang and Gu, Shuo and Yue, Yang and Huang, Gao and others},
  journal={arXiv preprint arXiv:2602.09023},
  year={2026}
}

@article{li2025gr,
  title={Gr-rl: Going dexterous and precise for long-horizon robotic manipulation},
  author={Li, Yunfei and Ma, Xiao and Xu, Jiafeng and Cui, Yu and Cui, Zhongren and Han, Zhigang and Huang, Liqun and Kong, Tao and Liu, Yuxiao and Niu, Hao and others},
  journal={arXiv preprint arXiv:2512.01801},
  year={2025}
}

@article{chen2025conrft,
  title={Conrft: A reinforced fine-tuning method for vla models via consistency policy},
  author={Chen, Yuhui and Tian, Shuai and Liu, Shugao and Zhou, Yingting and Li, Haoran and Zhao, Dongbin},
  journal={arXiv preprint arXiv:2502.05450},
  year={2025}
}

@article{liu2025mla,
  title={Mla: A multisensory language-action model for multimodal understanding and forecasting in robotic manipulation},
  author={Liu, Zhuoyang and Liu, Jiaming and Xu, Jiadong and Han, Nuowei and Gu, Chenyang and Chen, Hao and Zhou, Kaichen and Zhang, Renrui and Hsieh, Kai Chin and Wu, Kun and others},
  journal={arXiv preprint arXiv:2509.26642},
  year={2025}
}

@article{lin2025onetwovla,
  title={Onetwovla: A unified vision-language-action model with adaptive reasoning},
  author={Lin, Fanqi and Nai, Ruiqian and Hu, Yingdong and You, Jiacheng and Zhao, Junming and Gao, Yang},
  journal={arXiv preprint arXiv:2505.11917},
  year={2025}
}

@article{zawalski2024robotic,
  title={Robotic control via embodied chain-of-thought reasoning},
  author={Zawalski, Micha{\l} and Chen, William and Pertsch, Karl and Mees, Oier and Finn, Chelsea and Levine, Sergey},
  journal={arXiv preprint arXiv:2407.08693},
  year={2024}
}

@article{gu2025manualvla,
  title={ManualVLA: A Unified VLA Model for Chain-of-Thought Manual Generation and Robotic Manipulation},
  author={Gu, Chenyang and Liu, Jiaming and Chen, Hao and Huang, Runzhong and Wuwu, Qingpo and Liu, Zhuoyang and Li, Xiaoqi and Li, Ying and Zhang, Renrui and Jia, Peng and others},
  journal={arXiv preprint arXiv:2512.02013},
  year={2025}
}

@article{liu2024rdt,
  title={Rdt-1b: a diffusion foundation model for bimanual manipulation},
  author={Liu, Songming and Wu, Lingxuan and Li, Bangguo and Tan, Hengkai and Chen, Huayu and Wang, Zhengyi and Xu, Ke and Su, Hang and Zhu, Jun},
  journal={arXiv preprint arXiv:2410.07864},
  year={2024}
}

@article{tan2025interactive,
  title={Interactive post-training for vision-language-action models},
  author={Tan, Shuhan and Dou, Kairan and Zhao, Yue and Kr{\"a}henb{\"u}hl, Philipp},
  journal={arXiv preprint arXiv:2505.17016},
  year={2025}
}

@article{wen2024diffusion,
  title={Diffusion-vla: Generalizable and interpretable robot foundation model via self-generated reasoning},
  author={Wen, Junjie and Zhu, Minjie and Zhu, Yichen and Tang, Zhibin and Li, Jinming and Zhou, Zhongyi and Li, Chengmeng and Liu, Xiaoyu and Peng, Yaxin and Shen, Chaomin and others},
  journal={arXiv preprint arXiv:2412.03293},
  year={2024}
}

@article{zang2025rlinf,
  title={Rlinf-vla: A unified and efficient framework for vla+ rl training},
  author={Zang, Hongzhi and Wei, Mingjie and Xu, Si and Wu, Yongji and Guo, Zhen and Wang, Yuanqing and Lin, Hao and Shi, Liangzhi and Xie, Yuqing and Xu, Zhexuan and others},
  journal={arXiv preprint arXiv:2510.06710},
  year={2025}
}

@article{liu2026rdt2,
  title={RDT2: Exploring the Scaling Limit of UMI Data Towards Zero-Shot Cross-Embodiment Generalization},
  author={Liu, Songming and Li, Bangguo and Ma, Kai and Wu, Lingxuan and Tan, Hengkai and Ouyang, Xiao and Su, Hang and Zhu, Jun},
  journal={arXiv preprint arXiv:2602.03310},
  year={2026}
}

@article{pertsch2025fast,
  title={Fast: Efficient action tokenization for vision-language-action models},
  author={Pertsch, Karl and Stachowicz, Kyle and Ichter, Brian and Driess, Danny and Nair, Suraj and Vuong, Quan and Mees, Oier and Finn, Chelsea and Levine, Sergey},
  journal={arXiv preprint arXiv:2501.09747},
  year={2025}
}

@article{zhen20243d,
  title={3d-vla: A 3d vision-language-action generative world model},
  author={Zhen, Haoyu and Qiu, Xiaowen and Chen, Peihao and Yang, Jincheng and Yan, Xin and Du, Yilun and Hong, Yining and Gan, Chuang},
  journal={arXiv preprint arXiv:2403.09631},
  year={2024}
}

@article{chen2025reasoning,
  title={Reasoning Beyond Language: A Comprehensive Survey on Latent Chain-of-Thought Reasoning},
  author={Chen, Xinghao and Zhao, Anhao and Xia, Heming and Lu, Xuan and Wang, Hanlin and Chen, Yanjun and Zhang, Wei and Wang, Jian and Li, Wenjie and Shen, Xiaoyu},
  journal={arXiv preprint arXiv:2505.16782},
  year={2025}
}

@article{yang2025machine,
  title={Machine Mental Imagery: Empower Multimodal Reasoning with Latent Visual Tokens},
  author={Yang, Zeyuan and Yu, Xueyang and Chen, Delin and Shen, Maohao and Gan, Chuang},
  journal={arXiv preprint arXiv:2506.17218},
  year={2025}
}

@article{wang2025monet,
  title={Monet: Reasoning in Latent Visual Space Beyond Images and Language},
  author={Wang, Qixun and Shi, Yang and Wang, Yifei and Zhang, Yuanxing and Wan, Pengfei and Gai, Kun and Ying, Xianghua and Wang, Yisen},
  journal={arXiv preprint arXiv:2511.21395},
  year={2025}
}

@article{zhang2025reinbot,
  title={Reinbot: Amplifying robot visual-language manipulation with reinforcement learning},
  author={Zhang, Hongyin and Zhuang, Zifeng and Zhao, Han and Ding, Pengxiang and Lu, Hongchao and Wang, Donglin},
  journal={arXiv preprint arXiv:2505.07395},
  year={2025}
}

@misc{chen2025tgrpofinetuningvisionlanguageactionmodel,
      title={TGRPO :Fine-tuning Vision-Language-Action Model via Trajectory-wise Group Relative Policy Optimization}, 
      author={Zengjue Chen and Runliang Niu and He Kong and Qi Wang and Qianli Xing and Zipei Fan},
      year={2025},
      eprint={2506.08440},
      archivePrefix={arXiv},
      primaryClass={cs.RO},
      url={https://arxiv.org/abs/2506.08440}, 
}

@article{intelligence2026pi,
  title={$pi_{0.7}$: a Steerable Generalist Robotic Foundation Model with Emergent Capabilities},
  author={Intelligence, Physical and Ai, Bo and Amin, Ali and Aniceto, Raichelle and Balakrishna, Ashwin and Balke, Greg and Black, Kevin and Bokinsky, George and Cao, Shihao and Charbonnier, Thomas and others},
  journal={arXiv preprint arXiv:2604.15483},
  year={2026}
}

@inproceedings{zhai2023sigmoid,
  title={Sigmoid loss for language image pre-training},
  author={Zhai, Xiaohua and Mustafa, Basil and Kolesnikov, Alexander and Beyer, Lucas},
  booktitle={Proceedings of the IEEE/CVF international conference on computer vision},
  pages={11975--11986},
  year={2023}
}

%%%%%%%%%%%%%%%%%%%%%%%%%%%%%%%%%%%%%%%%%%%%%%%%%%%%%%%%%%%%

\newpage
\appendix

\section{Related Work}

\paragraph{Vision-Language-Action (VLA) Models.} 

Based on pretrained vision-language models (VLMs), VLA models~\cite{liu2025hybridvla, black2024pi_0, li2024cogact, liu2024rdt,luo2026look} have demonstrated promising performance in robotic control with discrete action generation~\cite{kim2024openvla, pertsch2025fast} or continuous action representation. Recent frameworks have diversified into regression-based~\cite{kim2025fine}, diffusion-based~\cite{wen2024diffusion, chen2025fast, gu2025manualvla, liu2024rdt}, and flow-matching-based architectures~\cite{bjorck2025gr00t, intelligence2025pi_, liu2026rdt2}. To further enhance precision and spatial awareness, several studies incorporate 3D spatial information or point-cloud reasoning into the VLA backbone~\cite{zhen20243d, qu2025spatialvla}, significantly boosting performance in complex manipulation tasks.
Inspired by Chain-of-Thought (CoT) reasoning in VLMs~\cite{wei2022chain}, recent research focuses on endowing VLAs with powerful reasoning capabilities. This includes generating intermediate text~\cite{intelligence2025pi_, lin2025onetwovla}, visual~\cite{cai2026internvla, zhao2025cot}, or multimodal plans~\cite{gu2025manualvla}, notably through methods like Embedded CoT~\cite{zawalski2024robotic} to strengthen spatial reasoning. However, these explicit decoding processes inevitably introduce inference latency. To alleviate this, further research explores introducing latent reasoning~\cite{chen2025reasoning, yang2025machine} into the action generation process~\cite{bai2026latent, liu2026last,cai2026internvla}, balancing performance gains with execution efficiency.
While these VLA frameworks have evolved significantly, the field remains predominantly confined to the imitation learning paradigm, where the lack of environmental interaction hinders the models' ability to generalize beyond static datasets or further refine their internal reasoning for robust control.

\paragraph{Reinforcement Learning (RL) for VLA Models.}

To overcome the limitations of static imitation learning, Reinforcement Learning (RL) has been increasingly adopted to fine-tune VLA models through environmental feedback \cite{tan2025interactive}. Early efforts focus on offline preference alignment \cite{zhang2025reinbot,zhang2024grape}, such as GRAPE~\cite{zhang2024grape}, which employs Direct Preference Optimization (DPO) \cite{rafailov2023direct} to align robot trajectories with human intent. To further enhance closed-loop robustness, online RL frameworks have emerged \cite{liu2025can,li2025simplevla,chen2025pirl,lu2025vla,xu2026twinrl,pan2026sop,chen2025conrft,zang2025rlinf}. For example, VLA-RL~\cite{lu2025vla} and RL4VLA~\cite{liu2025can} utilize Proximal Policy Optimization (PPO) \cite{schulman2017proximal} to improve generalization in unseen environments, while SimpleVLA-RL~\cite{li2025simplevla} and TGRPO~\cite{chen2025tgrpofinetuningvisionlanguageactionmodel} apply Group Relative Policy Optimization (GRPO) \cite{shao2024deepseekmath} to achieve efficient post-training without complex reward modeling.
Depending on the action space, these methods optimize either discrete token probabilities~\cite{li2025simplevla, lu2025vla} or continuous action sequences~\cite{chen2025pirl}. 
However, a critical gap remains: current RL-based VLA methods predominantly support vanilla architectures, focusing exclusively on action-space supervision. They fail to address how to conduct RL post-training for reasoning-based VLAs, leaving the intrinsic link between the internal "thought" process and the final "act" unoptimized.
Therefore, we present LaST-R1, a RL post-training framework with a unified VLA model that sequentially performs latent reasoning and action generation. Building on this, we propose Latent-to-Action Policy Optimization (LAPO) to concurrently optimize both the reasoning trajectory and the action generation, achieving superior robustness and precision in complex manipulation tasks.

\section{Derivation of the Latent Ratio}
\label{appendix:latent_likelihood_ratio}

In our Latent-to-Action Policy Optimization (LAPO) framework, we optimize both the action tokens and the continuous latent reasoning embeddings. While the likelihood ratio for action tokens is straightforward to compute via categorical probabilities, the ratio for the continuous latent embeddings requires defining a density function over the high-dimensional continuous space to estimate. Here, we provide the detailed mathematical derivation for the decision step-level latent ratio $r_t^z(\theta)$.

\textbf{1. Distribution Assumption}

Let $D$ denote the dimensionality of a single latent reasoning embedding. During policy optimization, we approximate the distribution of the generated continuous latents using an isotropic Gaussian distribution \cite{wang2025monet} centered at the current policy's output $\mathbf{z}_{t,i}^{\theta}$. We assume a fixed, predefined variance $\sigma^2$ across all dimensions. 
Thus, the probability density function (PDF) for a latent variable $\mathbf{x}$ generated by the current policy $\theta$ for the $i$-th latent embedding is given by:
\begin{equation}
    \pi_{\theta}(\mathbf{x}) = \frac{1}{(2\pi\sigma^2)^{D/2}} \exp\left( -\frac{1}{2\sigma^2} \|\mathbf{x} - \mathbf{z}_{t,i}^{\theta}\|^2 \right).
\end{equation}

\textbf{2. Evaluating the Numerator $\pi_{\theta}(\mathbf{z}_{t,i}^{\text{old}} \mid \cdot)$ and Denominator $\pi_{\theta_{\text{old}}}(\mathbf{z}_{t,i}^{\text{old}} \mid \cdot)$}

To compute the ratio for a single latent embedding sampled during a previous rollout, we must evaluate both the new policy $\theta$ and the old policy $\theta_{\text{old}}$ on the collected old latent embedding $\mathbf{z}_{t,i}^{\text{old}}$.

For the \textbf{numerator} $\pi_{\theta}(\mathbf{z}_{t,i}^{\text{old}} \mid \cdot)$ (the probability of the old latent embedding under the \textit{new} policy), we substitute $\mathbf{x} = \mathbf{z}_{t,i}^{\text{old}}$ into the PDF:
\begin{equation}
    \pi_{\theta}(\mathbf{z}_{t,i}^{\text{old}} \mid \cdot) = \frac{1}{(2\pi\sigma^2)^{D/2}} \exp\left( -\frac{1}{2\sigma^2} \|\mathbf{z}_{t,i}^{\text{old}} - \mathbf{z}_{t,i}^{\theta}\|^2 \right).
\end{equation}

For the \textbf{denominator} $\pi_{\theta_{\text{old}}}(\mathbf{z}_{t,i}^{\text{old}} \mid \cdot)$ (the probability of the old latent embedding under the \textit{old} policy), the mean of the old policy $\theta_{\text{old}}$ is, by definition, the sampled latent embedding $\mathbf{z}_{t,i}^{\text{old}}$. Consequently, the squared Euclidean distance becomes zero, i.e., $\|\mathbf{z}_{t,i}^{\text{old}} - \mathbf{z}_{t,i}^{\text{old}}\|^2 = 0$. The exponential term simplifies to $\exp(0) = 1$, leaving only the normalization constant:
\begin{equation}
    \pi_{\theta_{\text{old}}}(\mathbf{z}_{t,i}^{\text{old}} \mid \cdot) = \frac{1}{(2\pi\sigma^2)^{D/2}}.
\end{equation}

\textbf{3. Computing the Single Latent Embedding Ratio}

By dividing the numerator $\pi_{\theta}(\mathbf{z}_{t,i}^{\text{old}} \mid \cdot)$ by the denominator $\pi_{\theta_{\text{old}}}(\mathbf{z}_{t,i}^{\text{old}} \mid \cdot)$, the complex normalization constant $\frac{1}{(2\pi\sigma^2)^{D/2}}$ perfectly cancels out. The ratio for a single latent embedding $i$ is elegantly reduced to the negative exponentiated distance:
\begin{equation}
    r_{t,i}^{z}(\theta) = \frac{\pi_{\theta}(\mathbf{z}_{t,i}^{\text{old}} \mid \cdot)}{\pi_{\theta_{\text{old}}}(\mathbf{z}_{t,i}^{\text{old}} \mid \cdot)} = \exp\left( -\frac{1}{2\sigma^2} \|\mathbf{z}_{t,i}^{\text{old}} - \mathbf{z}_{t,i}^{\theta}\|^2 \right).
\end{equation}

\textbf{4. Sequence-Level Latent Reasoning Embeddings Aggregation}

The latent reasoning process consists of a sequence of $N_z$ latent embeddings. Assuming the continuous latents are conditionally independent given the rollout context, the joint ratio for the entire latent sequence $\mathbf{Z}_t$ is the product of the individual embedding ratios. Since the product of exponentials equals the exponential of their sum, we arrive at our final formulation:
\begin{equation}
    r_t^{z}(\theta) = \prod_{i=1}^{N_z} r_{t,i}^z(\theta) = \exp\left(-\frac{1}{2\sigma^2} \sum_{i=1}^{N_z} \| \mathbf{z}_{t,i}^{\text{old}} - \mathbf{z}_{t,i}^{\theta} \|^2 \right).
\end{equation}

\section{Dataset Details}

\subsection{Large Scale Pre-Training Datasets}
\label{appendix:pretrain}
To ensure LaST-R1 inherits a robust foundation of motor primitives and physical common sense, we curated a diverse corpus of 400K trajectories (28M frames) from the Open-X-Embodiment \cite{o2024open}, DROID \cite{khazatsky2024droid}, and RoboMIND \cite{wu2024robomind} repositories. Table \ref{tab:datasets} details the proportion of each dataset used in our pre-training mixture. Notably, beyond adhering to standard data quality filtering practices established by prior VLA works \cite{liu2026last}, we additionally ensure that all robot state annotations are accurate and physically consistent. Crucially, to empower the reasoning process without introducing computational bottlenecks during training, we precompute the DINOv3-based latent tokens for all pre-training frames offline. As detailed in Section \ref{method:architecture}, this involves extracting the \texttt{<CLS>} token from DINOv3 and applying top-$k$ selection to generate dense, semantically rich image representations. Because these spatial and temporal latent targets are processed entirely offline, they are simply loaded alongside the standard visual and textual inputs during this pretraining stage. This strategy provides the model with highly informative cognitive anchors to learn environmental dynamics and spatial reasoning, while adding virtually zero computational overhead to the pre-training pipeline. Ultimately, this stage establishes a shared representation space across diverse robotic datasets, enabling a seamless integration of reasoning and execution within the unified VLA framework.

\begin{table}[h]
\centering
\small
\renewcommand{\arraystretch}{1.1}
\setlength{\tabcolsep}{6pt}
\caption{\textbf{Datasets for Pre-training.}
The names of selected datasets for large-scale pretraining and their sampling ratios (\%).}
\vspace{-0.2cm}
\label{tab:datasets}
\begin{tabular}{lc | lc}
    \toprule
    \multicolumn{4}{c}{\textbf{Training Dataset Mixture}} \\
    \midrule
    \textbf{Dataset} & \textbf{Ratio (\%)} & \textbf{Dataset} & \textbf{Ratio (\%)} \\
    \midrule
    BridgeV2~\cite{ebert2022bridge,walke2023bridgedata} & 20.82 & Nyu Franka Play~\cite{cui2022play} & 0.24 \\
    Kuka~\cite{kalashnikov2018qt} & 20.22 & Stanford Hydra~\cite{belkhale2023hydra} & 0.20 \\
    Fractal~\cite{rt12022arxiv} & 13.67 & RoboMIND~\cite{wu2024robomind} & 0.20 \\
    Robo-Net~\cite{dasari2020robonet} & 11.53 & Jaco Play~\cite{dass2023jacoplay} & 0.19 \\
    Language Table~\cite{lynch2023interactive} & 7.72 & Dobb-E~\cite{shafiullah2023dobbe} & 0.18 \\
    BC-Z~\cite{jang2022bc} & 7.54 & Toto~\cite{zhou2023train} & 0.17 \\
    Maniskill~\cite{gu2023maniskill2unifiedbenchmarkgeneralizable} & 5.26 & Furniture Bench~\cite{heo2023furniturebench} & 0.09 \\
    DROID~\cite{khazatsky2024droid} & 4.82 & Utokyo Pr2 Tabletop~\cite{oh2023pr2utokyodatasets} & 0.04 \\
    Roboset~\cite{kumar2023robohive} & 3.21 & Utokyo Xarm Pap~\cite{matsushima2023weblab} & 0.04 \\
    FMB Dataset~\cite{luo2023fmb} & 1.50 & CMU Stretch~\cite{mendonca2023structured} & 0.02 \\
    Taco Play~\cite{rosetebeas2022latent,mees2023grounding} & 1.26 & DLR Sara Grid Clamp~\cite{padalkar2023guided} & 0.02 \\
    RoboTurk~\cite{DBLP:journals/corr/abs-1811-02790} & 0.70 & Utokyo Pr2 Fridge~\cite{oh2023pr2utokyodatasets} & 0.01 \\
    Berkeley Autolab Ur5~\cite{BerkeleyUR5Website} & 0.35 & --- & ---\\ 
    \bottomrule
\end{tabular}
\end{table}

\begin{figure*}[t]
    \centering
    % \vspace{-0.2cm}
    \includegraphics[width=\textwidth]{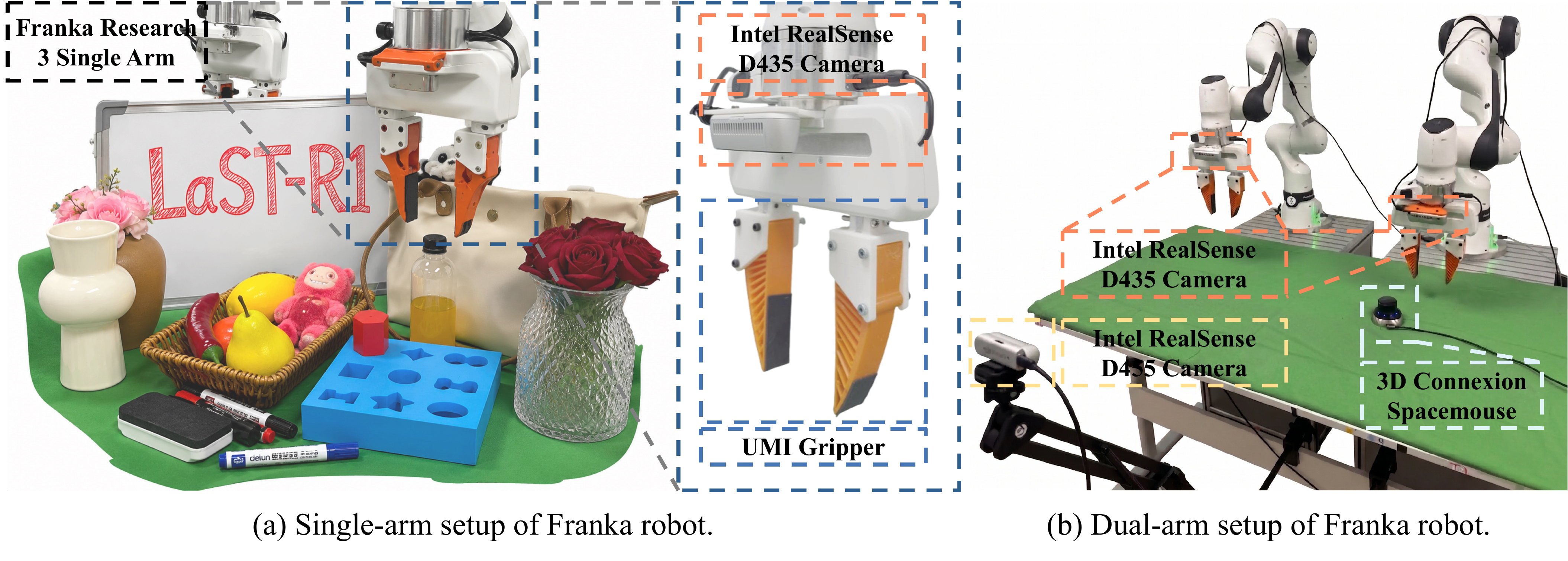}
    \caption{\textbf{Real-World Robot Experiment Setups of Franka Panda Arms.} }
    \label{fig:asset}
    \vspace{-0.3cm}
\end{figure*}

\subsection{Real-World Data Collection}
We configure the dual-arm experimental setup with two Franka Research 3 robotic arms equipped with 3D-printed UMI grippers. A single Intel RealSense D455 camera provides a third-person view, while two D435 cameras are mounted for wrist views. Inference is performed on an NVIDIA RTX 4090 GPU. Both SFT data collection and human-in-the-loop real-world RL are conducted using a 3D SpaceMouse. The main assets used throughout our experiments are shown in Figure \ref{fig:asset}.

\emph{1. Insert hexagon block.} This task requires the robot to hold  a hexagonal block and accurately insert it into a matching slot on a base. Success depends on the model's ability to perform precise 6-DoF spatial alignment and maintain accurate trajectory control to handle the tight physical clearance between the object and the target.

\emph{2. Open bag zipper.} The robot is required to use dual arms cooperatively, with one arm stabilizing a soft bag while the other grasps and pulls the zipper. This task evaluates the model's capacity for bimanual coordination and its ability to execute a smooth, continuous pulling trajectory while interacting with a deformable object.

\emph{3. Wipe vase with sponge.} This task requires the robot to hold a vase securely with one arm while using a sponge in the other to wipe its exterior. This demands highly accurate bimanual synchronization and the ability to predict and maintain continuous, stable contact along a curved 3D spatial trajectory.

\emph{4. Open bottle cap.} The robot must establish a stable grasp on a bottle with one arm while the other arm twists off its cap. This task evaluates the model's ability to execute fine motor control, specifically coordinating precise rotational manipulation and applying appropriate torque while simultaneously stabilizing the base object.

\section{Training Details}

\subsection{Alternative Latent Implementation Details}
\label{appendix:latent_detail}
To systematically evaluate the architectural design of our latent reasoning tokens, we compare our proposed DINOv3-based representation against three alternative compression strategies: \textbf{1) Global Pooling}, \textbf{2) Convolutional Downsampling}, and \textbf{3) a Lightweight Q-Former} \cite{li2023blip}. The most straightforward baseline, Global Pooling, trivially collapses the $N_v$ visual tokens from the VLA vision encoder into a single vector via token-dimensional average pooling, which operates without additional parameters but severely sacrifices fine-grained spatial details. To explicitly model the spatial hierarchy, the Convolutional baseline reshapes the $N_v$ visual tokens back into a 2D feature map and applies a sequence of $1 \times 1$ projections, aggressive strided downsampling (stride of 8), and global spatial convolutions with GELU activations to squeeze the map into a single flattened token. Alternatively, the Q-Former baseline employs an attention-based mechanism where a single learnable query adaptively aggregates task-relevant context from the dense $N_v$ visual tokens through stacked Pre-LayerNorm cross-attention and MLP blocks. While these baselines introduce varying degrees of parametric complexity to compress visual features, our proposed approach simply extracts the \texttt{<CLS>} token from a pre-trained DINOv3 model and applies top-$k$ dimension selection entirely offline. By directly harnessing the structurally rich and semantically dense feature space of a powerful vision foundation model, our DINOv3-based method circumvents the information loss inherent to naive pooling, spatial downsampling, or from-scratch query optimization. Consequently, this offline extraction strategy provides an optimal, computationally free cognitive anchor, ultimately achieving the best reasoning performance and highest success rates in our empirical evaluations.

\begin{figure*}[t]
    \centering
    % \vspace{-0.2cm}
    \includegraphics[width=0.5\textwidth]{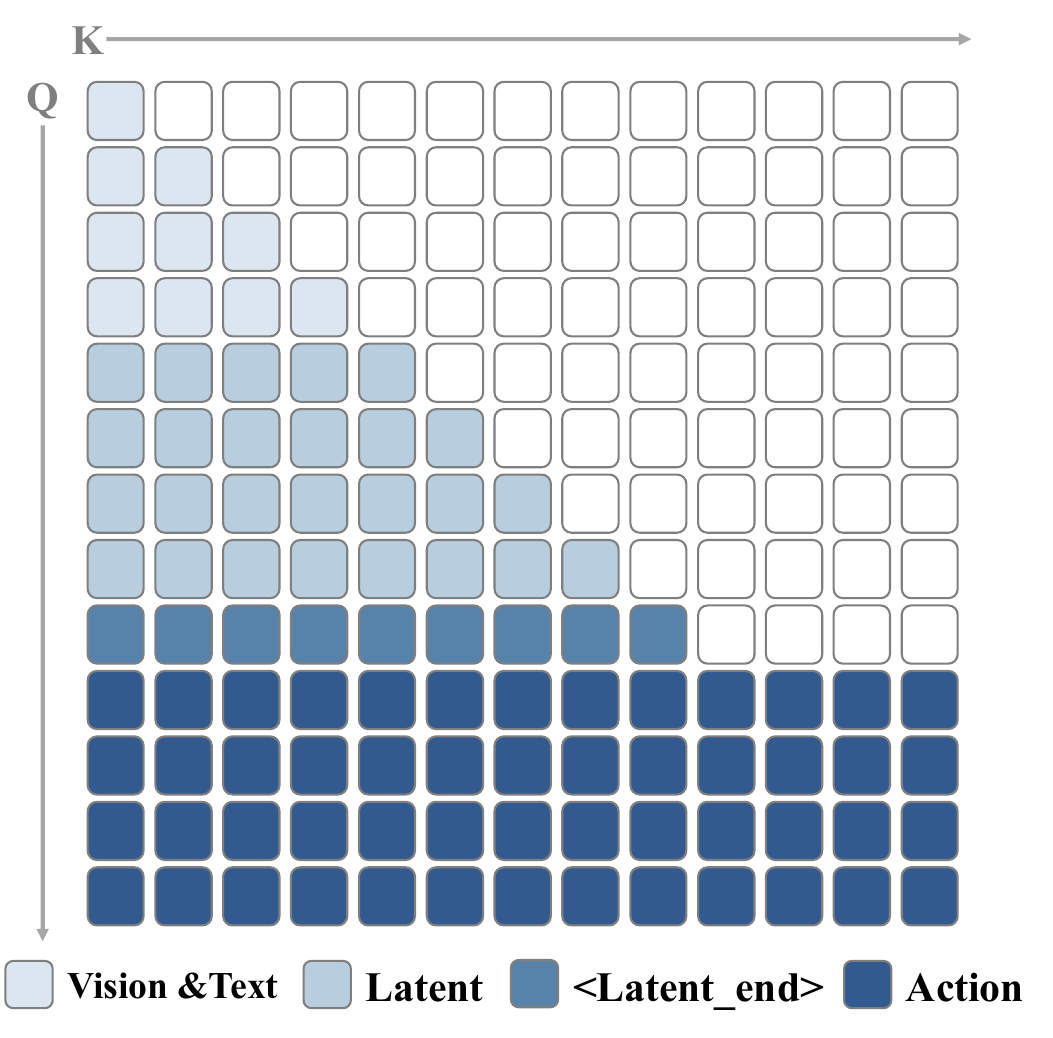}
    % \vspace{-0.6cm}
    \caption{\textbf{Hybrid Attention Mask Design.} Our model employs a custom attention mask to unify autoregressive reasoning and parallel action execution. The vision and text prompts, alongside the latent reasoning tokens, utilize a causal lower-triangular mask for sequential generation. After the \texttt{<Latent\_end>} transition token aggregates the full reasoning context, the action tokens employ a bidirectional mask. This allows all action tokens within a chunk to attend to the entire historical context as well as to each other, enabling efficient parallel decoding.}
    \label{fig:appd_mask}
    % \vspace{-0.3cm}
\end{figure*}

\subsection{Warm-up Details}
\label{appendix:warmup}

Our policy is initialized with pre-trained Qwen3-VL-4B \cite{bai2025qwen3vltechnicalreport} weights, expanding the tokenizer vocabulary to include discrete action tokens (\texttt{<action\_i>} for $i \in [0, 255]$) and a special transition token (\texttt{<latent\_end>}). As illustrated in Figure \ref{fig:appd_mask}, we adopt a hybrid decoding architecture: latent tokens are generated autoregressively to exploit the VLM's robust reasoning, while action tokens are decoded in parallel with a chunk size of 8 for physical execution efficiency. During training, we tailor the latent length strategy to the target domain. For the multi-task LIBERO benchmark, we encourage length exploration by uniformly sampling the reasoning length from $\{2, 4, 6, 8\}$ per sample (anchored on a maximum length of 8 with $M=4$ candidate positions); conversely, for single-task real-world deployments, we strictly enforce a fixed latent length of 8. The network is optimized using a joint objective comprising a cosine similarity loss for the predicted latents against offline ground-truth (GT) latents, a Cross-Entropy (CE) loss for the \texttt{<latent\_end>} token, and a standard CE loss for the parallel-decoded action tokens. These components are empirically weighted at $1 : 0.1 : 1$ to balance token-level gradients. Training is distributed across 8 $\times$ H20 GPUs using Accelerate \cite{accelerate} and DeepSpeed in \texttt{bf16} mixed precision, with a global batch size of 64 and the AdamW \cite{loshchilov2017decoupled} optimizer (peak learning rate $1 \times 10^{-5}$, cosine decay with a minimum ratio of 0.1). Finally, we customize the training duration per environment: the four LIBERO suites are trained in an extremely low-data regime (1 expert trajectory per task) for 10K iterations each, whereas real-world models are fine-tuned on 30 trajectories per task for 1K iterations.

\subsection{RL Training Details on LIBERO.} 
\label{appendix:rl_libero}
We implement our Latent-to-Action Policy Optimization (LAPO) framework using \texttt{verl} \cite{sheng2025hybridflow} with Ray \cite{moritz2018ray} and Fully Sharded Data Parallel (FSDP) on a single 8 $\times$ H20 GPU node. The VLA policy is initialized from the SFT warm-up checkpoint for continuous online interaction. During rollouts, the policy decodes an action chunk of 8 future action steps in parallel (yielding exactly 56 tokens per chunk) with a sampling temperature of 1.6, while trajectory lengths are capped at 240 steps for LIBERO-Spatial, 320 for LIBERO-Object/LIBERO-Goal, and 576 for LIBERO-Long. A strict verifier mechanism yields a sparse binary reward (0 or 1) for task success, which is scaled by a factor of 5 at the terminal step. Advantages are computed via GAE ($\gamma = 0.99$, $\lambda = 0.95$) with valid-timestep masking applied to accommodate variable-length trajectories. At each RL iteration, a rollout batch of 512 sampled trajectories is evenly partitioned into 4 mini-batches to perform 4 PPO optimization epochs. We train the policy using learning rates of $3 \times 10^{-5}$ for the VLA actor and $3 \times 10^{-4}$ for the value head. To ensure stability and efficiency, we apply global gradient clipping at 10, asymmetric PPO clipping bounds ($\epsilon_{\text{min}} = 0.2$, $\epsilon_{\text{max}} = 0.28$), and gradient/optimizer state offloading, alongside evaluating the model's success rate every 5 update steps.

\subsection{RL Training Details on Real-World}
\label{ap:Real-World RL}
Building upon the continuous asynchronous actor-learner pipeline \cite{luo2024serl,luo2025precise}, policy rollout and model optimization run concurrently without pausing data collection. 
The actor continuously executes the policy—which predicts actions conditioned on language prompts and dual-view RGB observations—and streams transition tuples to the learner. To preserve critical corrective signals, human interventions are routed to a dedicated buffer, allowing the learner to sample mixed mini-batches from both demonstration and rollout streams. We initialize the LaST-R1 model using the supervised warm-up checkpoint, and to ensure stable and computationally tractable online updates, we freeze the base model and exclusively update newly injected Low-Rank Adaptation (LoRA) \cite{hu2022lora} parameters with rank $r=32$. Online optimization commences after collecting an initial replay threshold of 500 transitions. The learner optimizes a joint objective combining behavior cloning ($\lambda_{\text{BC}}=1.0$) and Q-guided policy improvement ($\lambda_Q=0.5$) using the AdamW \cite{loshchilov2017decoupled} optimizer with a learning rate of $1 \times 10^{-5}$, zero weight decay, and gradient clipping at a maximum norm of 1.0. We apply a critic-to-actor update ratio of 2:1, where the critic utilizes one-step temporal-difference targets with a discount factor of $\gamma=0.98$ and target value parameters are softly updated at a rate of $\tau=0.005$. Gradient accumulation is set to 16 micro-steps with a micro-batch size of 2 per stream, yielding an effective batch size of approximately 32. Finally, to handle sparse real-world feedback, the environment provides a terminal reward of $+10$ upon operator-confirmed task completion, coupled with a constant step penalty of $-0.05$ to encourage execution efficiency, 
while the learner periodically broadcasts updated weights back to the actor for hot-loading.

\begin{figure*}[t]
    \centering
    % \vspace{-0.2cm}
    \includegraphics[width=\textwidth]{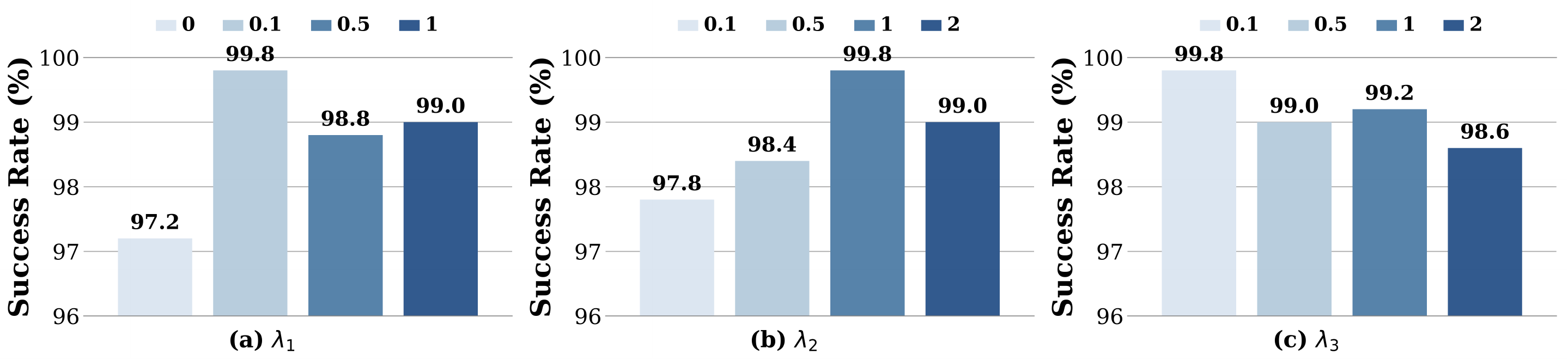}
    \vspace{-0.6cm}
    \caption{\textbf{Ablation Studies on Loss Coefficients.} Performance impact of varying (a) latent loss weight $\lambda_1$, (b) state-value weight $\lambda_2$, and (c) transition penalty $\lambda_3$. In each sub-figure, unlisted coefficients are fixed at their optimal values. The combination ($\lambda_1=0.1, \lambda_2=1, \lambda_3=0.1$) achieves the optimal gradient balance and highest success rate.}
    \label{fig:appd_lambda}
    % \vspace{-0.3cm}
\end{figure*}

\section{Additional Quantitative Analysis}

\subsection{Additional Ablation Studies on Hyperparameters}
\label{appendix:ablation}
To systematically evaluate the impact of the weighting hyperparameters in our joint training objective $\mathcal{L}_{\text{total}}(\theta)$, we conduct an ablation study varying the coefficients $\lambda_1$, $\lambda_2$, and $\lambda_3$ on LIBERO-Spatial. As shown in Figure \ref{fig:appd_lambda}, in each experiment, we isolate the effect of one coefficient while fixing the others at their empirically optimal values (i.e., $\lambda_1=0.1$, $\lambda_2=1$, $\lambda_3=0.1$). First, analyzing the latent loss weight $\lambda_1$, we observe that relying entirely on implicit gradient backpropagation from the action loss ($\lambda_1=0$) yields a suboptimal success rate of 97.2\%. Introducing explicit latent supervision significantly improves performance, peaking at 99.8\% with $\lambda_1=0.1$. However, excessively high values (e.g., $\lambda_1=1$) slightly degrade performance to 99.0\%, likely because the latent objective begins to overshadow the primary physical execution task. Second, for the state-value estimation weight $\lambda_2$, a balanced weight of $\lambda_2=1$ achieves the best result (99.8\%). Lower weights such as 0.1 and 0.5 degrade the success rate to 97.8\% and 98.4\% respectively, underscoring the necessity of robust value estimation for accurate advantage computation in our LAPO framework. Finally, evaluating the transition weight $\lambda_3$ reveals that a modest penalty of $\lambda_3=0.1$ is optimal (99.8\%). Increasing this weight up to 2 forces a sharp performance drop to 98.6\%, indicating that over-penalizing the \texttt{<latent\_end>} token disrupts the delicate exploration balance required for the policy to learn state-conditional reasoning lengths. Consequently, the configuration of $\lambda_1=0.1$, $\lambda_2=1$, and $\lambda_3=0.1$ establishes the effective balance across action generation, internal reasoning, and value estimation.

\begin{figure*}[t]
    \centering
    % \vspace{-0.2cm}
    \includegraphics[width=\textwidth]{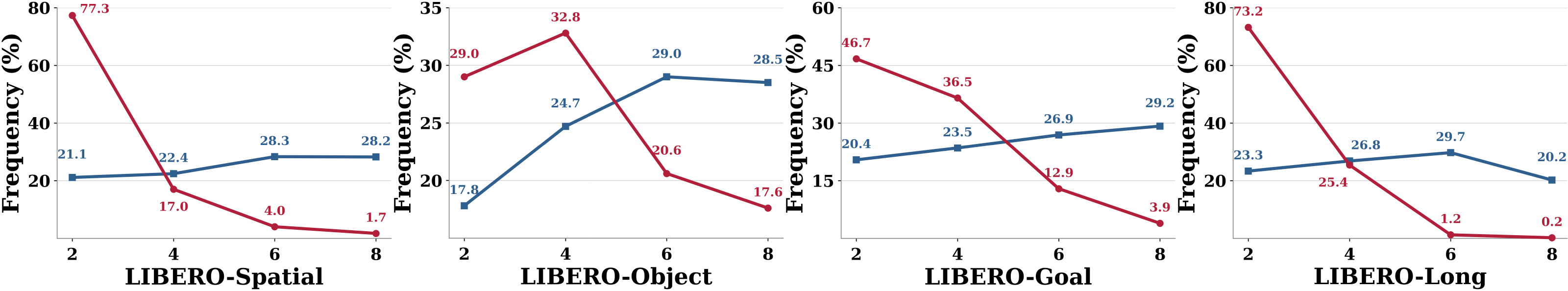} 
    % \vspace{-0.6cm}
    \caption{\textbf{Frequency Distribution of Adaptive Latent Reasoning Lengths.} Compared to the SFT warm-up (\textcolor[HTML]{2F5F8F}{blue}), the RL-optimized policy (\textcolor[HTML]{B11F3A}{red}) successfully learns the optimal number of latent reasoning steps, achieving an excellent balance between reasoning precision and inference efficiency.}
    \label{fig:appendix_frequency}
    % \vspace{-0.2cm}
\end{figure*}

\subsection{Analysis of Adaptive Reasoning Length}
\label{appd:adaptive_length}
To investigate the efficacy of our dynamic reasoning mechanism, we analyze the frequency distribution of the chosen latent reasoning lengths (bounded by a maximum of 8 tokens) across the four LIBERO task suites. Figure \ref{fig:appendix_frequency} illustrates the contrast between the policy right after SFT warm-up and the fully RL-optimized policy. During the warm-up phase, where reasoning lengths are sampled randomly, the pre-RL model exhibits a relatively uniform distribution across lengths 2, 4, 6, and 8. This indicates a lack of cognitive prioritization, treating simple and complex states with equal computational budgets. Following Latent-to-Action Policy Optimization (LAPO), the length distribution undergoes a dramatic shift. Guided by the environment reward and our transition-specific optimization objective, the model effectively learns an "early-exit" strategy. Across all environments, the RL-optimized policy heavily gravitates toward shorter cognitive horizons, with reasoning lengths of 2 or 4 tokens dominating the decision steps. This empirical behavior strongly validates our design: rather than passively executing fixed-length reasoning, LaST-R1 proactively learns to terminate its internal deliberation once sufficient state representations are formed. By drastically reducing the token count per step without compromising the near-perfect success rates, the model achieves an optimal balance between robust physical reasoning and inference efficiency.

\begin{figure*}[t]
    \centering
    % \vspace{-0.2cm}
    \includegraphics[width=0.8\textwidth]{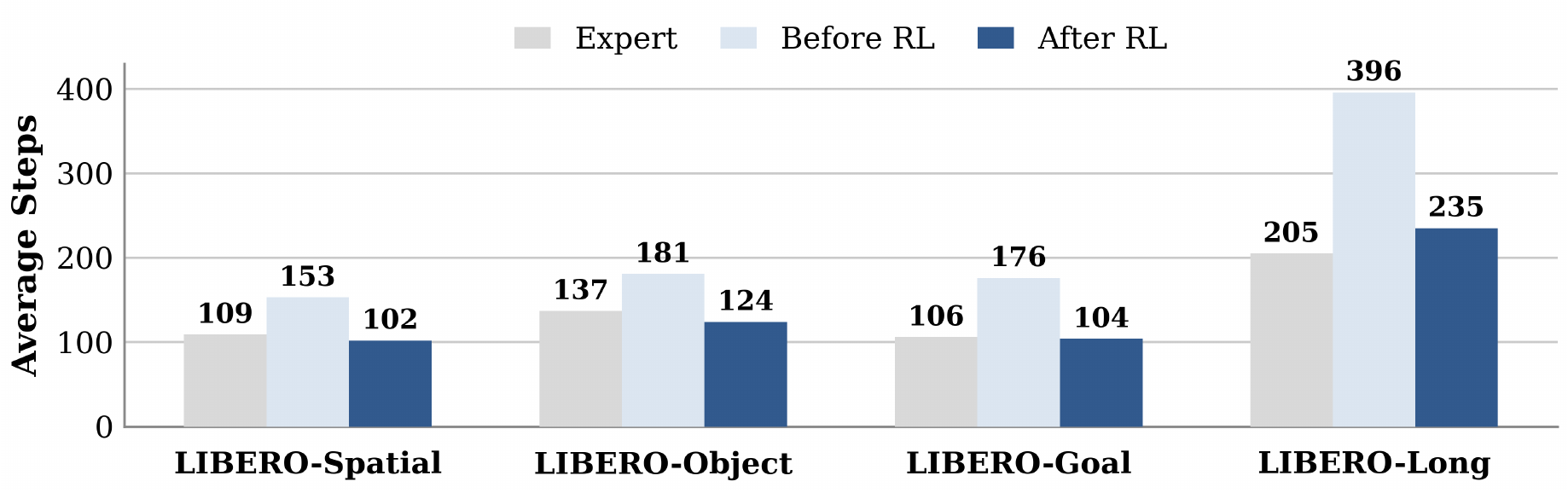} 
    % \vspace{-0.2cm}
    \caption{\textbf{Comparison of Average Execution Steps Across LIBERO Task Suites.} We report the average steps taken by the expert demonstrations, the pre-RL policy, and the RL-optimized policy. After optimized via LAPO, LaST-R1 not only drastically reduces the execution steps compared to the imitation baseline, but even surpasses the temporal efficiency of experts in most scenarios.}
    \label{fig:appendix_steps}
    % \vspace{-0.2cm}
\end{figure*}

\begin{figure*}[t]
    \centering
    % \vspace{-0.2cm}
    \includegraphics[width=\textwidth]{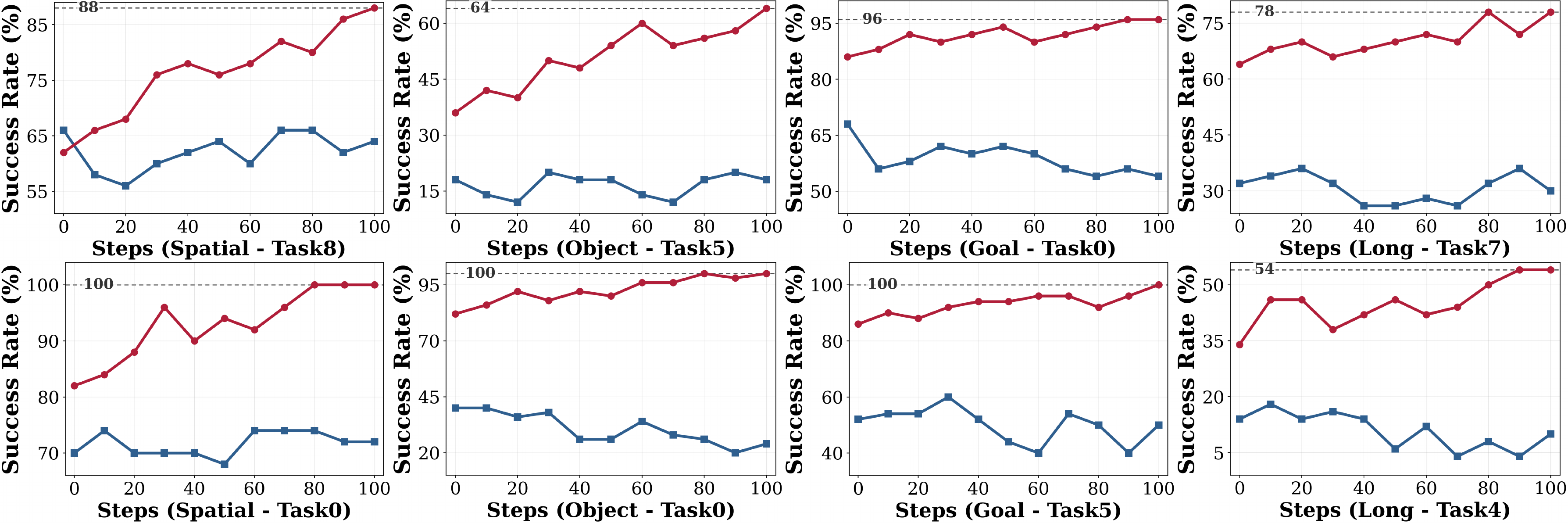}
    % \vspace{-0.6cm}
    \caption{\textbf{Generalization Analysis on LIBERO.} For each task suite, models are warmed up with one trajectory per task, followed by online RL training on 9 tasks, with the remaining 1 task for evaluation. While the out-of-distribution performance of the Action-Only PPO baseline (\textcolor[HTML]{2F5F8F}{blue}) stagnates, our LaST-R1 with LAPO (\textcolor[HTML]{B11F3A}{red}) shows continuous generalization improvements across all task suites.}
    \label{fig:appendix_gen}
    % \vspace{-0.2cm}
\end{figure*}

\subsection{Episode Length Comparisons}
\label{appd:episode_length}
To further demonstrate the efficacy of our framework, we evaluate the average number of execution steps across 500 test trajectories for each LIBERO suite. Notably, these averages are computed unconditionally over all rollouts, regardless of ultimate task success or failure. As illustrated in Figure \ref{fig:appendix_steps}, the policy trained after warm-up SFT (Before RL) requires significantly more steps than the simulator-provided expert demonstrations. This inflation in trajectory length stems from compounding errors and purely reactive behaviors; when the pre-RL model encounters out-of-distribution (OOD) states, it often gets trapped in inefficient loops or prolonged, unsuccessful wandering that approaches the maximum step limit.
However, following online interaction, the RL-optimized policy (After RL) exhibits a dramatic reduction in execution steps, notably outperforming even the original expert trajectories in the Spatial, Object, and Goal task suites. We attribute this exceptional temporal efficiency directly to the optimized latent reasoning process. Rather than passively mimicking the scripted expert data, which often relies on conservative motion planning and rigid, waypoint-based trajectories, the RL-driven reasoning strategy empowers the policy with a deeper, more generalized understanding of the task dynamics. By jointly optimizing the reasoning and action spaces, the model learns to synthesize highly efficient internal representations, thereby discovering optimal shortcuts and formulating more direct, precise, and decisive trajectories than the original algorithmic experts.

\subsection{Additional Generalization Analysis}
\label{ap:AGA}
To provide a more granular understanding of the out-of-distribution (OOD) learning process, Figure \ref{fig:appendix_gen} details the step-by-step generalization curves for two representative held-out tasks from each LIBERO suite. The learning dynamics reveal a stark contrast between the two paradigms throughout the online interaction phase.
The Action-Only PPO baseline (blue) exhibits a classic overfitting pathology. Across almost all tasks, its OOD success rate flatlines early in training and frequently oscillates or even collapses (e.g., Spatial-Task8 and Long-Task7). This empirical evidence confirms that optimizing purely in the action space forces the model to memorize the specific kinematic trajectories of the 9 training tasks, leaving it entirely brittle and incapable of adapting when faced with the unseen spatial configurations or objects of the held-out task.
In sharp contrast, our LaST-R1 optimized with LAPO (red) displays a remarkably stable and continuous upward trajectory across the entire training process. The curves demonstrate no signs of catastrophic forgetting or OOD performance degradation. Notably, our policy not only rapidly converges to a perfect 100\% OOD success rate on several tasks (Spatial-Task0, Object-Task0, and Goal-Task5), but also shows robust, monotonic growth on the most challenging multi-stage tasks. For instance, in Long-Task4, while the action-only baseline completely fails to surpass a 20\% success rate due to compounding errors, our method steadily climbs to 54\%. This extended analysis clearly illustrates that jointly optimizing the latent reasoning space empowers the policy to abstract high-level semantic and physical principles, allowing it to dynamically compose learned skills for novel scenarios rather than rigidly overfitting to the training distribution.

\section{Additional Qualitative Analysis}

\subsection{Action-to-Vision Attention}
\label{appd:a2v attn}

\begin{figure*}[t]
    \centering
    % \vspace{-0.2cm}
    \includegraphics[width=0.9\textwidth]{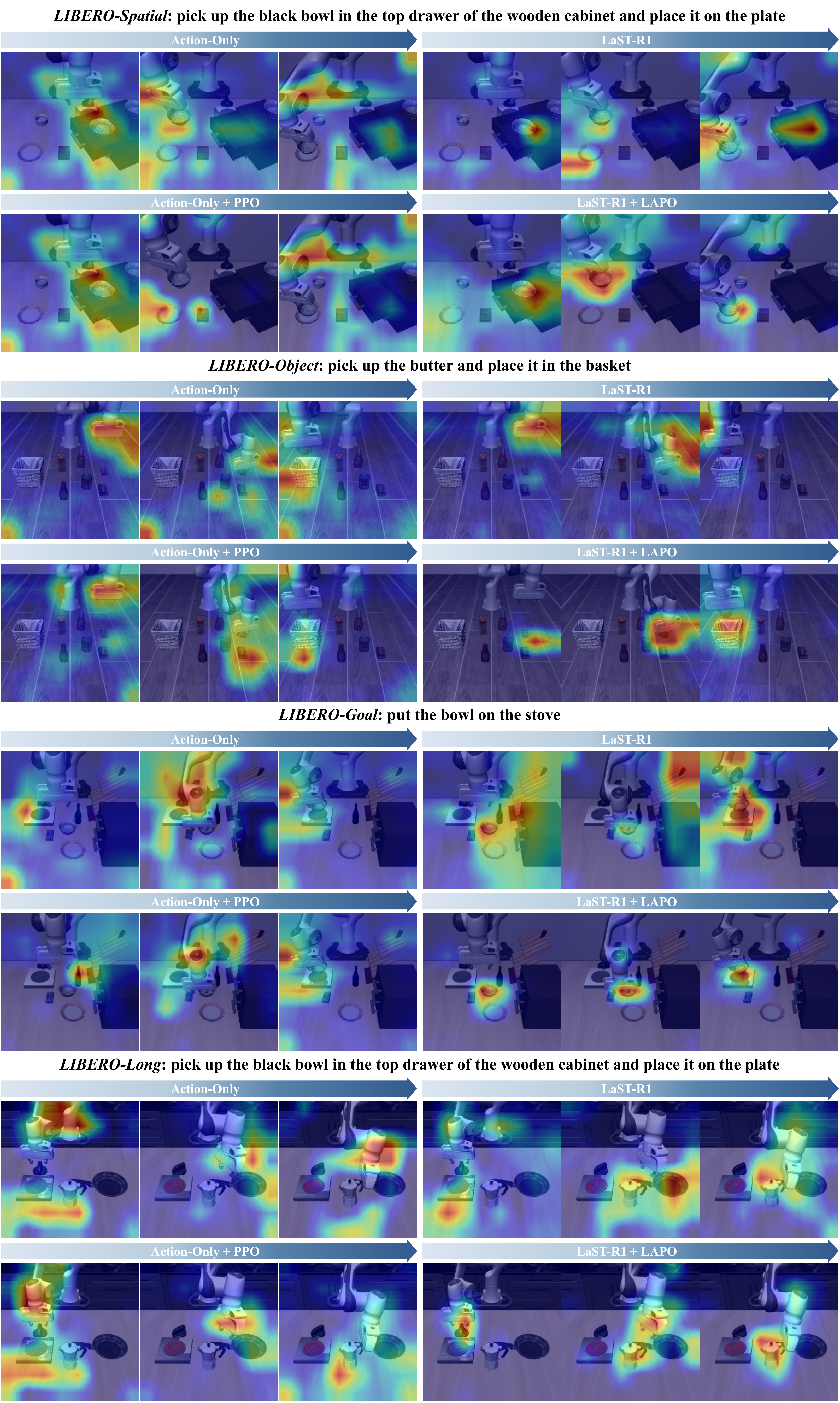} 
    % \vspace{-0.6cm}
    \caption{\textbf{Visualizations of Action-to-Vision Cross-Attention.} We compare the attention maps across four representative trajectories from four task suites in LIBERO. }
    \label{fig:appendix_attn}
    % \vspace{-0.2cm}
\end{figure*}

To gain deeper insights into the internal decision-making mechanisms of our policy, we utilize Grad-CAM \cite{selvaraju2017grad} to visualize the cross-attention weights from the action tokens to the visual tokens. As illustrated in Figure \ref{fig:appendix_attn}, we compare spatial attention maps across four representative trajectories from the LIBERO task suites to evaluate the impact of explicit latent reasoning and our LAPO framework. Examining the SFT-trained models, the Action-Only policy frequently exhibits diffused and scattered attention, struggling to consistently localize task-relevant entities and often lagging behind the robot's end-effector. In stark contrast, by generating latent reasoning tokens prior to action decoding, our proposed LaST-R1 establishes a strong semantic anchor, resulting in highly concentrated and precise visual grounding on critical objects of interaction. These advantages become even more pronounced after reinforcement learning post-training. While applying standard PPO to the Action-Only model tends to over-focus on the gripper's immediate vicinity and lacks long-horizon awareness, LaST-R1 + LAPO achieves the most robust, goal-directed attention. Because LAPO jointly optimizes the latent reasoning space and the action space using the environment reward, the policy learns to dynamically shift its intense visual focus from the manipulated object to the target receptacle as the trajectory progresses, confirming that our framework successfully aligns internal visual cognition with precise physical execution.

\begin{figure*}[t]
    \centering
    % \vspace{-0.2cm}
    \includegraphics[width=\textwidth]{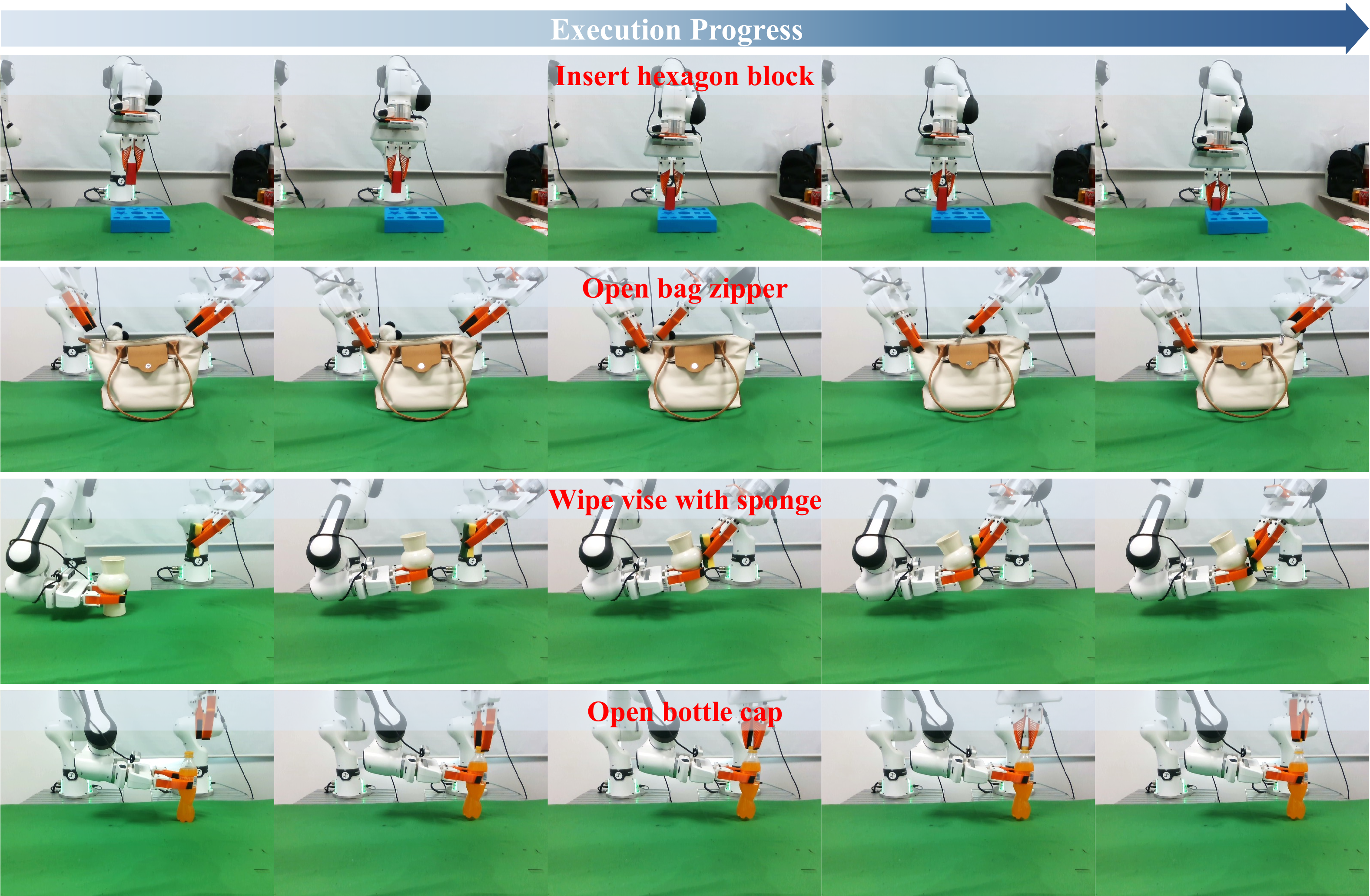} 
    % \vspace{-0.3cm}
    \caption{\textbf{Real-World Execution Trajectories of the Proposed Policy.} The sequences illustrate the continuous execution progress across diverse robotic manipulation tasks.}
    \label{fig:appendix_real_rollout}
    % \vspace{-0.3cm}
\end{figure*}

\begin{figure*}[t]
    \centering
    % \vspace{-0.2cm}
    \includegraphics[width=\textwidth]{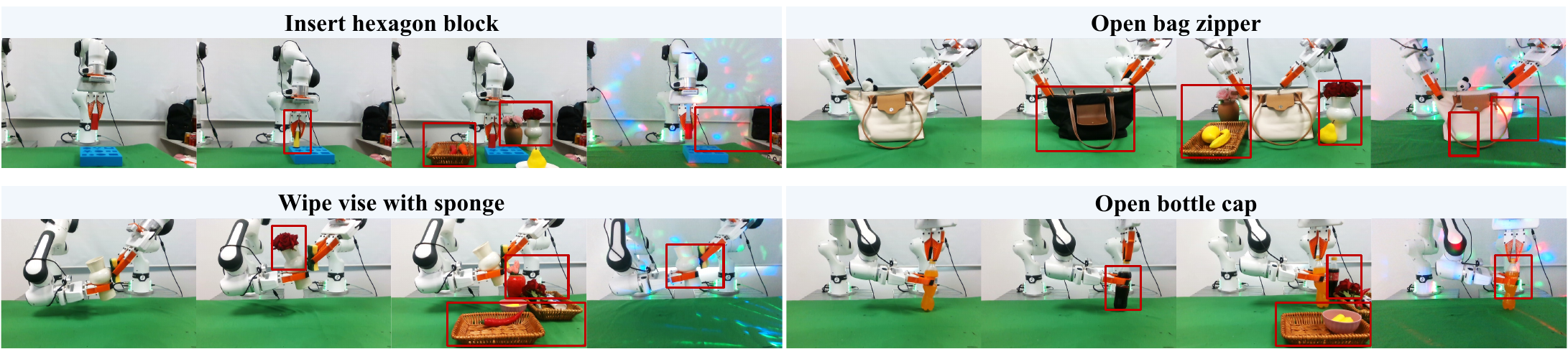} 
    % \vspace{-0.6cm}
    \caption{\textbf{Policy Robustness Under Unseen Configurations.} The model consistently maintains task success when subjected to severe visual variations, effectively ignoring out-of-distribution distractors, unseen object instances, and dynamic lighting conditions (indicated by red bounding boxes).}
    \label{fig:appendix_ood_rollout}
    % \vspace{-0.2cm}
\end{figure*}

\begin{figure*}[t]
    \centering
    % \vspace{-0.2cm}
    \includegraphics[width=\textwidth]{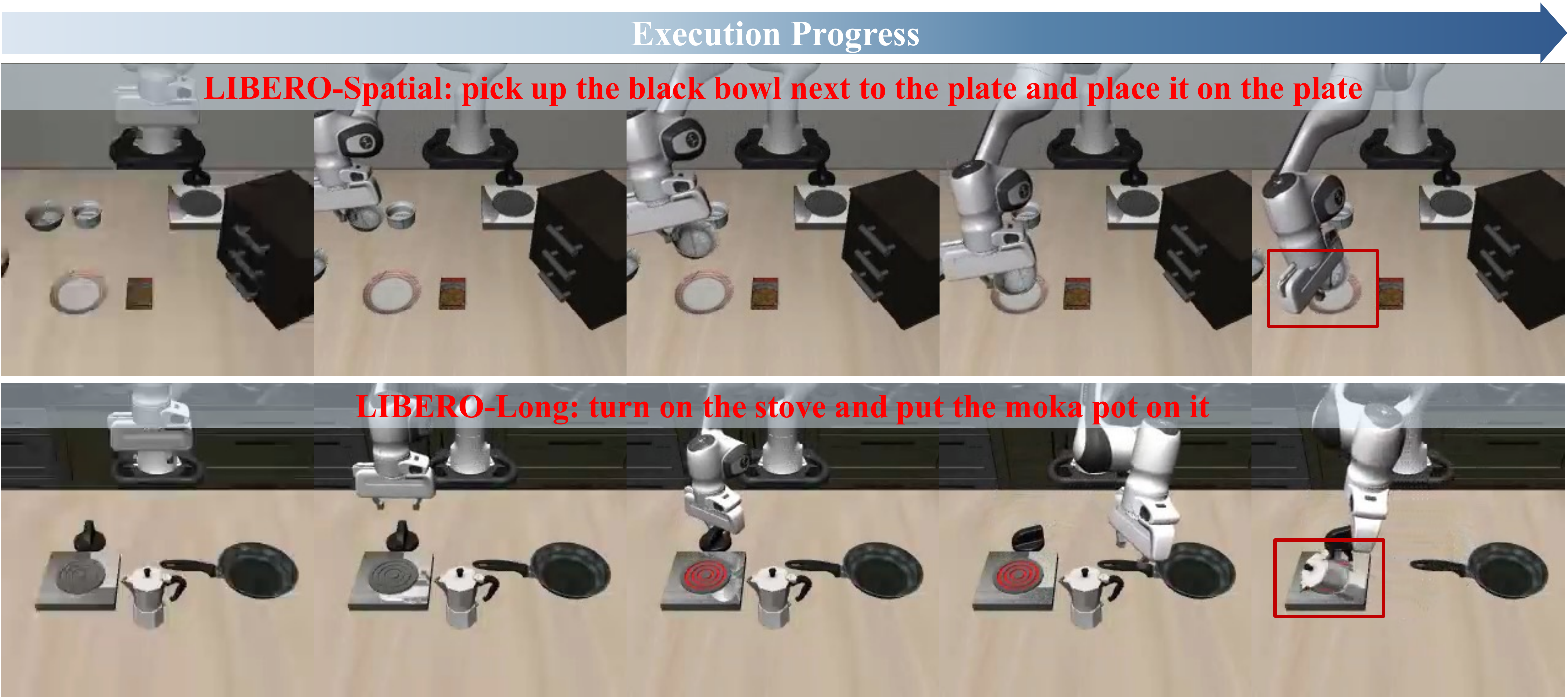} 
    % \vspace{-0.6cm}
    \caption{\textbf{Failure Cases in the LIBERO.} Red boxes highlight the moments of failure.}
    \label{fig:appendix_failure_sim}
    % \vspace{-0.2cm}
\end{figure*}

\begin{figure*}[t]
    \centering
    % \vspace{-0.2cm}
    \includegraphics[width=\textwidth]{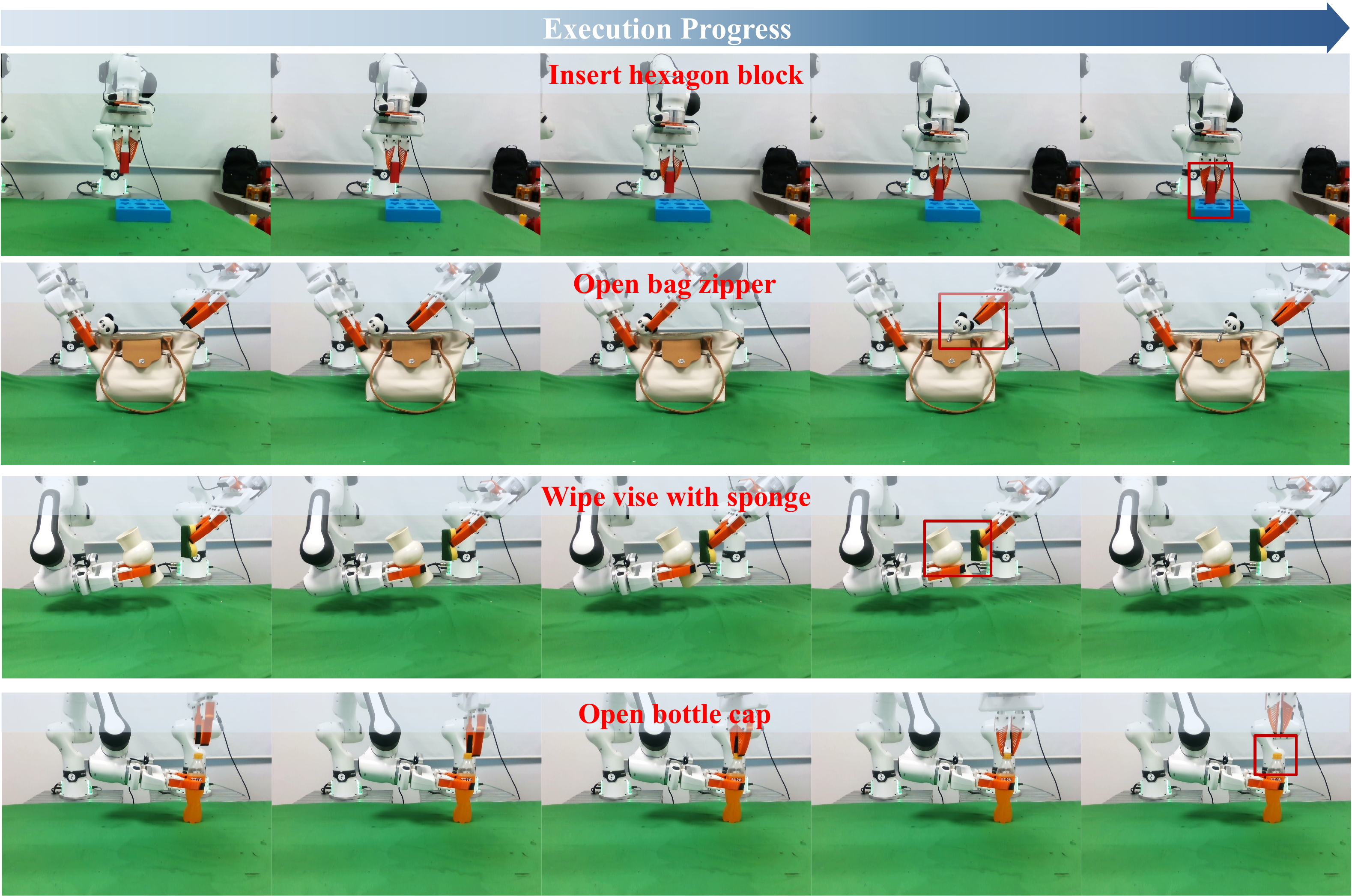} 
    % \vspace{-0.6cm}
    \caption{\textbf{Visualizations of Real-World Failure Cases.} The sequences illustrate typical execution errors across four tasks, with red bounding boxes highlighting the specific moments of failure.}
    \label{fig:appendix_failure_real}
    % \vspace{-0.2cm}
\end{figure*}

\subsection{Real-World Visualization}
\label{appd:real_v}
To evaluate the practical deployment capabilities of LaST-R1, we visualize its execution trajectories across diverse real-world manipulation tasks, as shown in Figure \ref{fig:appendix_real_rollout}. As demonstrated, the policy exhibits stable closed-loop control during standard execution (e.g., precise insertion, bimanual coordination). Crucially, the model demonstrates remarkable robustness in visually cluttered environments, as shown in Figure \ref{fig:appendix_ood_rollout}. Despite severe out-of-distribution perturbations—such as dynamic lighting, novel distractor objects, and background variations—LaST-R1 consistently maintains execution stability, highlighting its strong generalization capabilities in complex, unconstrained scenarios.

\section{Failure Case Analysis}

\subsection{Simulation.} In the rare failure cases observed, the model's errors stem from fine-grained physical execution. As shown in Figure \ref{fig:appendix_failure_sim}, in the LIBERO-Spatial task: \emph{pick up the black bowl next to the plate and place it on the plate}, the robot successfully grasps the bowl but releases it slightly off-center, causing it to slip off the edge of the plate due to imperfect placement precision. Similarly, in the LIBERO-Long task: \emph{turn on the stove and put the moka pot on it}, while the robot correctly turns on the stove, its arm inadvertently knocks over the moka pot during the approach phase. This irreversible change in the object's pose prevents subsequent grasping, highlighting the inherent long-tail challenge of maintaining strictly collision-free trajectory planning in complex, multi-step environments.

\subsection{Real-World.} 
In real-world deployments, the observed failure cases primarily stem from fine-grained execution imperfections and complex contact dynamics. As shown in Figure \ref{fig:appendix_failure_real}, for the \emph{Insert hexagon block} task, the failure is caused by a slight positional deviation during the final insertion phase, preventing the block from fitting into the slot. In the \emph{Open bag zipper} task, the robot struggles with dynamic friction, causing the gripper to slip and detach from the zipper mid-pull. For the \emph{Wipe vase with sponge} task, the failure occurs because the robot fails to establish actual physical contact between the sponge and the vase surface, resulting in an ineffective wiping motion. Finally, in the \emph{Open bottle cap} task, the robot successfully grasps and twists the cap but terminates the action prematurely, failing to unscrew the cap completely. These cases indicate that although LaST-R1 achieves remarkable performance in handling continuous, high-precision physical feedback in real-world contact-rich scenarios, a certain probability of failure still exists.

\section{Limitations and Future Work}
While LaST-R1, optimized via the proposed LAPO framework, exhibits exceptionally strong performance across both the LIBERO benchmark and real-world deployments, it has several limitations that present exciting avenues for future research. First, to strike an optimal balance between reasoning capacity and inference speed, our current adaptive CoT mechanism caps the maximum latent sequence length at 8 and restricts the early-exit transitions to a predefined set of $M$ candidate positions. Moving forward, we aim to refine the latent representation mechanism and explore more flexible, fully dynamic reasoning architectures that can halt deliberation without manually designed constraints. Second, our current real-world evaluations are conducted exclusively on single-arm and dual-arm manipulation platforms. To demonstrate the scalability and universality of our framework, we plan to deploy LaST-R1 to more complex robotic embodiments, extending our latent reasoning paradigm to multi-fingered dexterous hands and whole-body humanoid robot control.

\section{Broader Impact}
Our work proposes LaST-R1 VLA, a foundation model for robotic manipulation that seamlessly integrates adaptive-length latent reasoning and parallel action execution within a unified framework. While our method significantly enhances spatial awareness and dynamically allocates cognitive compute through Latent-to-Action Policy Optimization (LAPO), deploying high-capacity VLA models in physical environments inherently introduces potential risks. These risks primarily involve unpredictable physical behaviors under out-of-distribution scenarios or misinterpretation of ambiguous human instructions.
To mitigate these challenges, future real-world deployments should incorporate rigorous safety guardrails and strict hardware-level operational constraints.
Broadly, our post-training framework delivers strong performance across diverse downstream tasks, facilitating the transition of robotic systems from controlled laboratory settings to real-world deployment. Meanwhile, by enabling robots to fluidly balance deliberate cognitive planning with rapid physical reflexes, our approach advances the development of reliable, general-purpose robotic assistants for complex, unstructured domains such as smart manufacturing, elder care, and domestic automation.

%%%%%%%%%%%%%%%%%%%%%%%%%%%%%%%%%%%%%%%%%%%%%%%%%%%%%%%%%%%%

\end{document}